\documentclass[10pt,twocolumn,letterpaper]{article}

\usepackage{cvpr}
\usepackage{times}
\usepackage{epsfig}
\usepackage{graphicx}
\usepackage{amsmath}
\usepackage{amssymb}

\usepackage{url}
\usepackage{float}
\usepackage{subcaption}

\usepackage{xspace}
\usepackage{array}

\usepackage[pagebackref=true,breaklinks=true,letterpaper=true,colorlinks,bookmarks=false]{hyperref}

\cvprfinalcopy 

\def\cvprPaperID{2083} 

\newcommand{\mscoco}{MS-COCO\xspace}
\newcommand{\mscocosub}{MS-COCO-sub\xspace}
\newcommand{\clutmnist}{MNIST-Single\xspace}
\newcommand{\multimnist}{MNIST-Multi\xspace}
\newcommand{\clutrender}{3D-Single\xspace}
\newcommand{\clutrenderinst}{3D-Single-Inst\xspace}
\newcommand{\multirender}{3D-Multi\xspace}
\newcommand{\multirenderinst}{3D-Multi-Inst\xspace}
\newcommand{\compfull}{\textsc{comp-full}\xspace}
\newcommand{\compobjonly}{\textsc{comp-obj-only}\xspace}
\newcommand{\compnomask}{\textsc{comp-no-mask}\xspace}

\newcommand{\baselineaug}{\textsc{baseline-aug}\xspace}
\newcommand{\baseline}{\textsc{baseline}\xspace}
\newcommand{\baselineaugreg}{\textsc{baseline-aug-reg}\xspace}
\newcommand{\baselinereg}{\textsc{baseline-reg}\xspace}
\ifcvprfinal\fi
\begin{document}

\title{Teaching Compositionality to CNNs\thanks{Preprint appearing in CVPR 2017.}}

\author{Austin Stone\\
  Yi Liu \and
  Huayan Wang\\
  D. Scott Phoenix \and
  Michael Stark\\
  Dileep George \and
Vicarious FPC, San Francisco, CA, USA\\
{\tt\small \{austin, huayan, michael, yi, scott, dileep\}@vicarious.com}
}

\maketitle

\begin{abstract}
Convolutional neural networks (CNNs) have shown great success in
computer vision, approaching human-level performance when trained for
specific tasks via application-specific loss functions.
In this paper, we propose a 
method for augmenting and training
CNNs so that their learned features are 
compositional. 
It encourages networks to form representations that 
disentangle objects from their surroundings and from each other, thereby promoting
better generalization. 
%
Our method is agnostic to the specific details of the underlying CNN
to which it is applied and can in principle be used with any CNN.
%
As we show in our experiments, the learned representations lead to
feature activations that are more localized and improve
performance over non-compositional baselines in object recognition
tasks.

\end{abstract}

%
\section{Introduction}
\label{sec:intro}
%
%
%
%
%
Convolutional neural networks (CNNs) have shown remarkable performance
in many computer vision tasks~\cite{lecun1989backpropagation,
ImageNetCNN,GoogleNet,Zisserman15,CNNOverview} including image
classification~\cite{ImageNetCNN}, object class
detection~\cite{szegedy13nips,hariharan14eccv}, instance
segmentation~\cite{hariharan15cvpr}, image
captioning~\cite{karpathy15cvpr,vinyals15icmlw}, and scene
understanding~\cite{eslami16nips}.
Their success is typically attributed to two factors; they have large
enough capacity to make effective use of the ever-increasing amount of
image training data available today, while at the same time managing
the number of free parameters through the use of inductive biases from
neuroscience. Specifically, the interleaving of locally connected
filter and pooling layers~\cite{HandW} bears similarity to the visual cortex's
interleaving of simple cells, which have localized receptive fields, and complex 
cells, which have wider receptive fields and greater local invariance.

\begin{figure}
\centering
\includegraphics[width=.99\columnwidth]{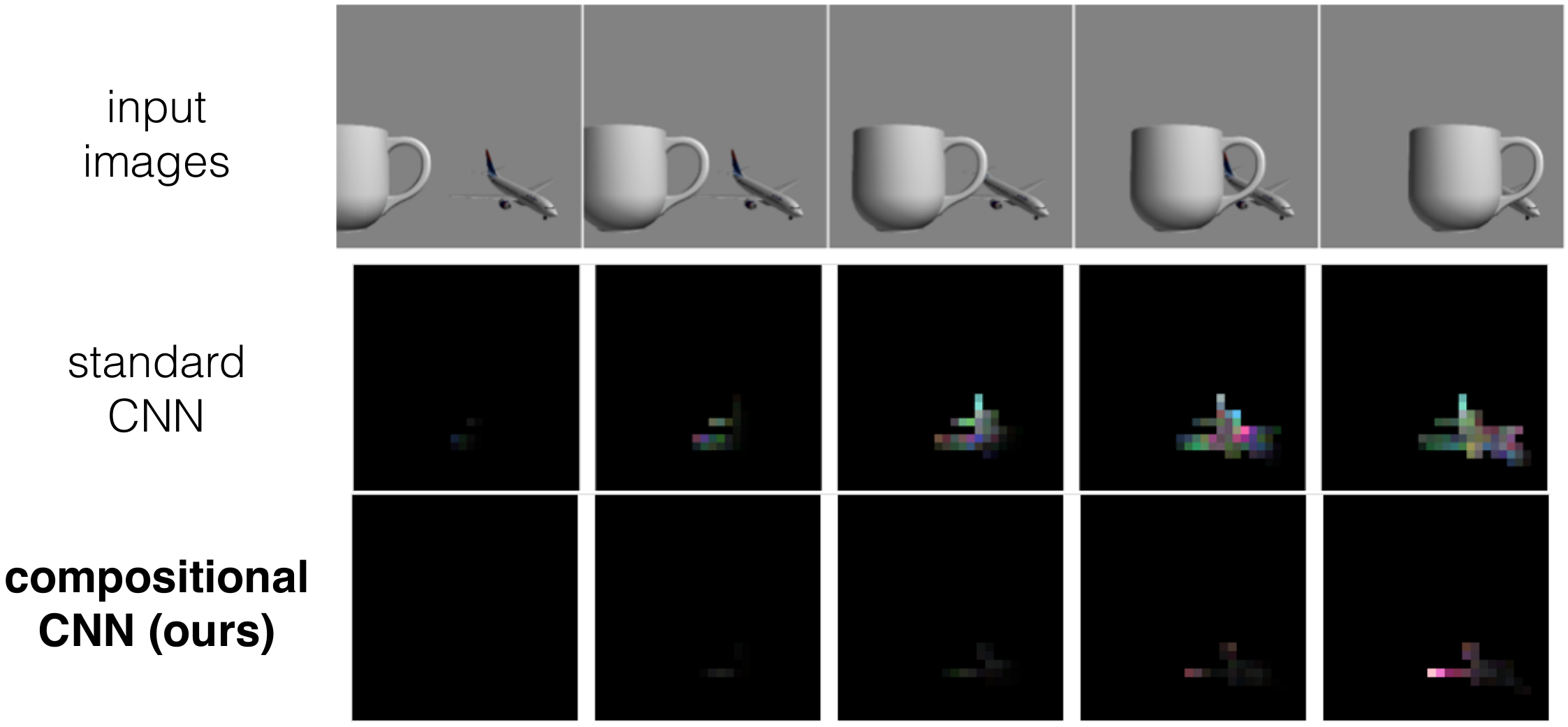} 
\caption{For a standard CNN (VGG, \cite{Zisserman15}), the presence of
a nearby object (cup) greatly affects the activations in the region of
an object of interest (airplane).
In contrast, a CNN trained with our method demonstrates
better compositionality in its feature representations -- the
activations in the airplane region represent primarily the airplane
and are therefore less affected by the presence of the
cup.}
\label{fig:VGGNotComp}
\end{figure}

Recently, researchers have investigated more inductive biases
from neuroscience to improve CNN
architectures. Examples include learning representations from video
sequences~\cite{agrawal15iccv,gao16accv,jayaraman16cvpr}, encouraging
the utilization of depth information~\cite{hoffman16cvpr}, and using
physical interaction with the environment~\cite{pinto16eccv} to bias
representations.

In this paper, we follow a similar philosophy, but focus our attention
on the inductive bias of {\em compositionality}: the notion that {\em the
representation of the whole should be composed of the representation
of its parts} (we give a
precise formal definition of this notion in Sect.~\ref{sec:approach}).
Intuitively, encouraging this property during training results in
representations that are more robust to re-combination (e.g., when
seeing a familiar object in a novel context) and less prone to
focusing on discriminative but irrelevant background features.
It is also in line with findings from neuroscience that suggest separate
processing of figure and ground regions in the 
visual cortex~\cite{BGSuppression2,BGSuppression}.
%
%
Note that a typical CNN does not exhibit this property
(Fig.~\ref{fig:VGGNotComp} visualizes the difference in
activations between a CNN trained without (VGG~\cite{Zisserman15}) and
with our compositionality objective\footnote{Fig.~\ref{fig:VGGNotComp}
shows the activation difference in the
airplane region between the current frame and a frame where the
airplane is shown in isolation. 
Activations are taken from intermediate conv. layers with spatial
resolution $28\times28$. We marginalize over feature channels to create
visualization.}).

In contrast to previous work that designs compositional
representations from the ground
up~\cite{salakhutdinov13pami,wang15cvpr,tabernik16arxiv,yuille16jmlr},
our approach does not mandate any particular network architecture or
parameterization -- instead, it comes in the form of a modified
training objective that can be applied to {\em teach any standard CNN}
about compositionality in a soft manner.
%
While our current implementation requires object masks for
training, 
it allows
apples to apples comparison of networks trained with or
without the compositionality objective. As our experiments show
(Sect.~\ref{sec:experiments}), the objective consistently improves
performance over non-compositional baselines.

This paper makes the following specific contributions:
First, we introduce a novel notion of {\em compositionality} as
  an inductive bias for training arbitrary convolutional neural
  networks (CNNs). It captures the intuition that {\em the
    representation of a partial image} should be equal to {\em the
    partial representation} of that image.
Second, we implement that notion in the form of a
  modified CNN training objective, which we show to be
  straightforward to optimize yet effective in learning compositional
  representations.
Third, we give an extensive experimental evaluation on
  both synthetic and real-world images that highlights the efficacy of
  our approach for object recognition tasks
  and demonstrates the contributions of different components of our
  objective.
\section{Related work}
\label{sec:related}
Our work is related mostly to three major lines of research:
compositional models, inductive biases, and the role of context in
visual recognition. 
\paragraph{Compositional models.}
Compositional models have existed since the early
days of computer vision~\cite{marr82vision} and have appeared mainly in two different varieties. 
The first flavor focuses on
the creation of hierarchical feature representations by means of
statistical modeling~\cite{fidler07cvpr,ommer10pami,zhu10cvpra,yuille16jmlr},
re-usable deformable parts~\cite{zhu10cvprb,ott11cvpr}, or
compositional graph structures~\cite{si13pami,wang15cvpr}.
The second flavor designs neural network-based representations 
in the form of recursive neural
networks~\cite{socher11icml}, imposing hierarchical priors on Deep
Boltzman Machines~\cite{salakhutdinov13pami}, or introducing parametric
network units that are themselves compositional~\cite{tabernik16arxiv}.

The basis for our work is a notion of compositionality
(Sect.~\ref{sec:soft-comp}) that is distinct from all these
approaches in that it does not have to be baked into the design of a
model but can be applied as a soft constraint to a CNN. 
Recent work~\cite{oquab15cvpr} constrains CNN activations to lie
within object masks in the context of weakly-supervised
localization. Our compositional objective (Sect.~\ref{sec:training})
goes beyond this formulation: it consists of multiple components
that not only suppress background activations, but also explicitly
encourage object activations to be invariant to both background
clutter and adjacent objects. Our experiments verify that each
component is important for performance
(Sect.~\ref{sec:exp_diagnostic}).
%
\paragraph{Inductive biases.}
A recent line of work on neural network architectures takes
inspiration from human learning in its design of training
regimen. 
It has demonstrated improved performance when training from video
sequences instead of still
images~\cite{agrawal15iccv,jayaraman16cvpr}, assuming an
object-centric view~\cite{gao16accv}, integrating multimodal sensory
side information~\cite{hoffman16cvpr}, or even being in control of
movement~\cite{pinto16eccv}. The benefit arises from
providing helpful inductive biases to the learner that regularize
the learned representations.
The inductive bias of compositionality presented in this work
(Sect.~\ref{sec:soft-comp}) follows a similar motivation but is
largely complementary to the biases explored by these prior
approaches.
%
%
\paragraph{The role of context in visual recognition.}
It is well known that context plays a major role in visual
recognition, both in human and artificial vision
systems~\cite{galleguillos07cviu,oliva07cs,divvala09cvpr}. Our
environment tends to be highly regular, and making use of regularities
in the occurrence of different object and scene classes has been shown
to be beneficial for recognizing familiar
objects~\cite{murphy03nips,cinbis12eccv}, objects in unusual
circumstances~\cite{choi12prl}, and recurring spatial
configurations~\cite{farhadi11cvpr,pepik13cvpr,gupta15arxiv,wu16pami}.
At the extreme, object classes can be successfully recognized even in
the absence of local information by relying exclusively on scene
context~\cite{russell07nips}.
%
%

While CNN-based representations typically support the use of context
implicitly (by including pixels indiscriminately in
a receptive field), they lack the ability to explicitly address
context and non-context information. The notion of compositionality
proposed in this work (Sect.~\ref{sec:soft-comp}) is a step
towards making CNN-based representations more amenable to
explicit context modeling through an external mechanism (by cleanly
separating the representation of objects from their context).
%
The experiments in this paper (Sect.~\ref{sec:experiments}) do not
further elaborate on this aspect, but indicate that the compositional
objective {\em (i)} elicits a performance improvement, {\em (ii)} the
improvement is similar for objects appearing in and out-of-context,
and {\em (iii)} the improvement is least pronounced for very
small object instances.
\section{Teaching compositionality to CNNs}
\label{sec:approach}
\begin{figure}[t]
  \centering
      \includegraphics[width=.95\columnwidth]{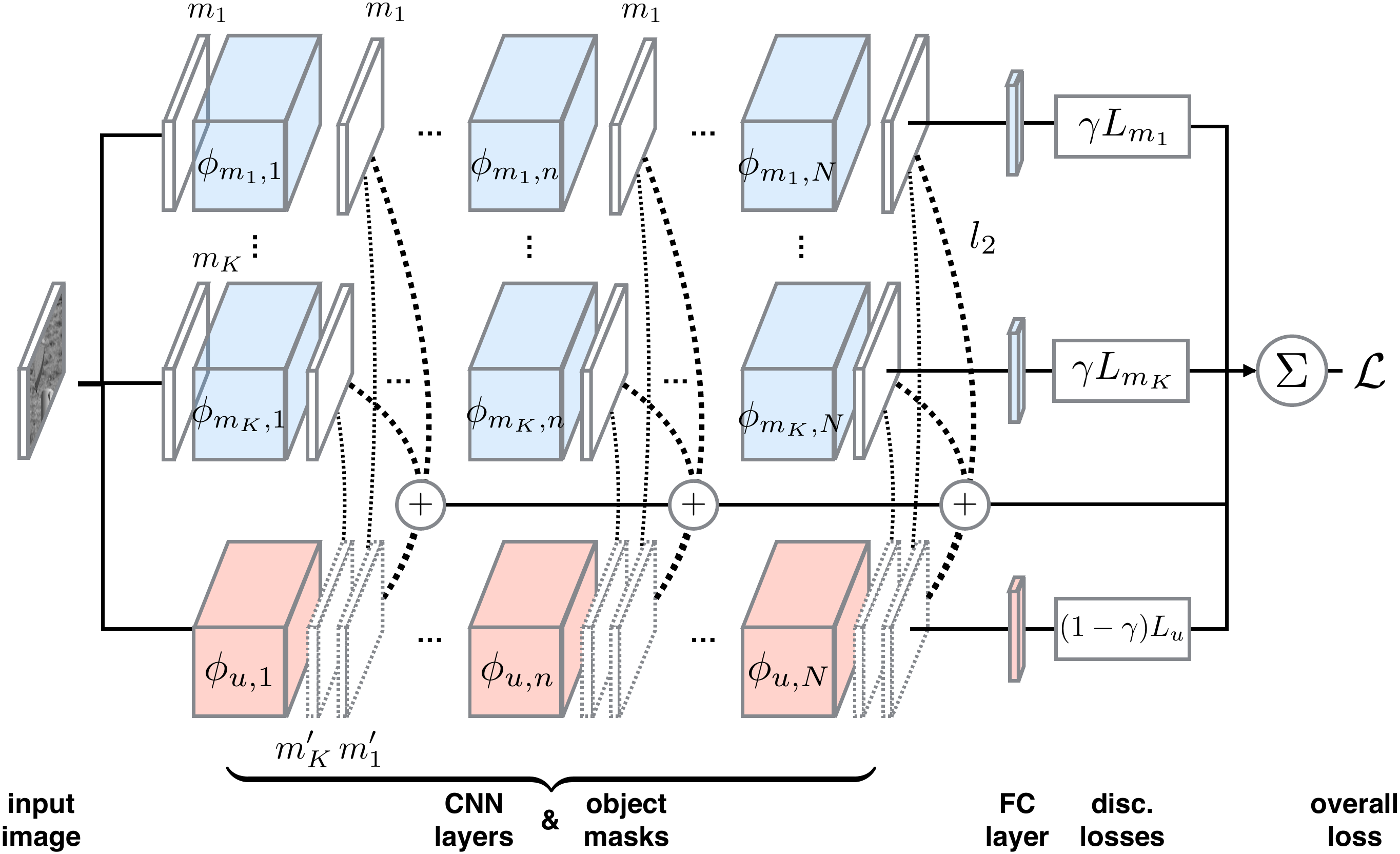}
    \caption{Architecture and loss function ($\mathcal{L}$)
      computation for encouraging compositionality when multiple
      objects are present in a training image (see
      Sect.~\ref{sec:architecture} and Sect.~\ref{sec:training}).
      The original CNN (red)
      is enhanced by $K$ additional masked CNNs (blue), all of them
      sharing weights.
      $\phi_{.,.}$ represent feature maps, $L_.$ loss functions, and
      $\gamma$ is a hyper-parameter. Masks $m_k$ (solid) are applied to
      feature map outputs, while masks $m'_k$ (dotted) are only
      applied for computing losses.
      For simplicity, we depict layers and masks with equal sizes.
}
\label{fig:arch}
\end{figure}
This section describes our approach to encouraging CNNs to learn
compositional representations. To that end, we proceed from
introducing our notion of compositionality
(Sect.~\ref{sec:soft-comp}) to describing network architecture
(Sect.~\ref{sec:architecture}) and training procedure
(Sect.~\ref{sec:training}) to giving technical details of our
implementation (Sect.~\ref{sec:details}).

\subsection{Compositionality notion} 
\label{sec:soft-comp}
The goal of our notion of compositionality is to encourage
the representation of a part of an image to be similar to the
corresponding part of the representation of that image.
More formally, let $X$ be an image, $m$ a binary mask that identifies
part of $X$ (i.e, $m$ is a tensor of the same shape as $X$ with $1$s
indicating part affiliation), $\phi$ a mapping from an image onto an
arbitrary feature layer of a CNN, and $p$ the projection operator
onto the feature map represented by $\phi$. We define $\phi$ to
be {\em compositional} iff the following holds:
\begin{equation}
\label{eq:comp}
\phi(m \cdot X) = p(m) \cdot \phi(X).
\end{equation}
%
Here, the $\cdot$ operator represents element-wise multiplication. The
projection operator, $p$, down-samples the object mask to the size of
the output of $\phi$. 
%
E.g., if $\phi(X)$ is the activations of a
convolutional layer with size $(h, w, c)$ (the first two
dimensions are spatial and $c$ is the number of feature channels), $p$
will down-sample the object mask to size $(h,w)$ and then stack $c$
copies of the down-sized object mask on top of each other to produce a
mask of size $(h, w, c)$.

Note that in practice we do not require Eq.~\eqref{eq:comp}
to hold for all possible masks $m$, as this would constrain
$\phi$ to be the identity map. Instead, we apply the inductive bias
selectively to image parts that we would like to be treated as a unit
-- obvious choices for these selected parts include objects or object parts.
In the following, we use object masks
(as provided by standard data sets such as~\mscoco~\cite{lin14eccv}) as
the basis for compositionality.

\subsection{Enhanced network architecture}
\label{sec:architecture}
To encourage a network to satisfy the compositionality
property of Eq.~\eqref{eq:comp} (Sect.~\ref{sec:soft-comp}), we devise
an enhanced architecture and corresponding objective function. Note that
this enhancement is non-destructive in nature and leaves the
original network completely intact; it merely makes virtual copies of
the original network, Fig.~\ref{fig:arch}.

When there is only one object in the input image,
teaching compositionality takes the form of ensuring that the activations
within the region of that object remain invariant regardless of what
background the object appears on.
With multiple objects, we also explicitly
ensure that the activations of each object remain the same as if that
object were shown in isolation (i.e., activations should be
invariant to the other objects within the respective object
mask).

To implement this notion, we create $K + 1$ weight-sharing CNNs where
$K$ is the number of objects shown in the scene. $K$ of these CNNs take as input
a different object instance, each shown against a blank background (we apply
the mask for the $k$\textit{th} object instance to the input image
before giving the input image to the $k$\textit{th} CNN). We refer to
these $K$ CNNs as ``masked CNNs," and we denote the mapping onto layer
$n$ of the $k$\textit{th} masked CNN as $\phi_{m_{k},n}$.

Each of these $K$ masked CNNs have their respective object mask
reapplied to their activations at multiple layers in the hierarchy
(see Sect.~\ref{sec:details}), zeroing out activations outside of the
object region. These masked activations are then passed on to higher
layers (which might also re-apply the mask again in the same way). This
constrains the masked CNNs to only use activations within the object
mask region when classifying the input image.
The final ($K + 1$\textit{th}) CNN receives as input the original
image with no masks applied, and we refer to it subsequently as the
``unmasked CNN". We denote the mapping onto layer 
$n$ of this CNN as $\phi_{u, n}$. We denote the total number of layers as $N$.

\subsection{Training procedure}
\label{sec:training}
We train the architecture of Sect.~\ref{sec:architecture} for
compositionality by introducing an objective function that combines an
application-specific discriminative CNN loss with additional
terms that establish dependencies between the different masked and
unmasked CNNs.
\paragraph{Discriminative loss.}
To encourage correct discrimination, we add separate
discriminative loss terms for both the $K$ masked and
the one unmasked CNN, denoted $L_{m_{k}}$ and $L_{u}$,
respectively. 
Their relative contributions are controlled by the
hyperparameter $\gamma \in [0, 1]$, to yield
\begin{equation} \label{eq:L_d}
\mathcal{L}_{d} = \frac{1}{K} \big(\sum_{k} \gamma L_{m_{k}}\big) + (1 - \gamma) L_{u}.
\end{equation}

\paragraph{Compositional loss.}
To encourage compositionality, we add $K \times N$ terms that
establish dependencies between the responses of corresponding layers
of the masked and unmasked CNNs, respectively.
Specifically, on all layers at which an object mask is applied, we
take the $l_2$ 
difference between the activations of the masked CNN and the 
activations of the unmasked CNN. We then multiply this difference by a
layer specific penalty hyper-parameter (denoted as $\lambda_{n}$) and
add this to our compositional loss:
\begin{equation} \label{eq:L_c}
\mathcal{L}_{c} = \frac{1}{K} \sum_{k} \sum_{n} \lambda_{n} \vert\vert \phi_{m_{k},n} - \phi_{u,n} m'_{k}\vert\vert_{2}^{2}.
\end{equation}
The final objective can then be stated simply as $\mathcal{L} =
\mathcal{L}_{d} + \mathcal{L}_{c}$.
%
Because the unmasked CNN sees all
objects and will naturally have different activations from the
$k$\textit{th} masked CNN due to the presence of the objects other
than the $k$th object, we apply a mask to the unmasked CNN's
activations before computing the penalty term. We denote this mask as
$m'_{k}$. However, we do not pass these masked activations ($\phi_{u,
  n} m'_{k}$) up to higher layers as was done for the masked CNNs; we
only use them to compute the compositional penalty
term on layer $n$.
%
\paragraph{Design choices.}
The above objective leaves degrees of freedom w.r.t. choosing
the precise nature of the masks $m'_{k}$, and the corresponding
choices do have an impact on performance
(Sect.~\ref{sec:exp_diagnostic}).
First, to penalize background activations outside of the regions of
objects of interest, we can make
$m'_{k}$ be a tensor of $1$s but with the locations of all objects
other than the $k$\textit{th} object filled with $0$s.
Second, we can penalize any shifts in activations within
the region of the $k$\textit{th} object without discouraging background
activations by making $m'_{k}$
equal to $m_{k}$.

\subsection{Implementation details}
\label{sec:details}
Our experiments (Sect.~\ref{sec:experiments}) use the following
network architectures:
{\footnotesize
\mscocosub (Sect.~\ref{sec:exp_mscoco}):
conv1-conv3 ($224 \times 224 \times 64$), pool1, conv4-conv6 ($128
\times 128 \times 128$), pool2, conv7-conv9 ($64 \times 64 \times
256$), pool3, conv10-conv12 ($32 \times 32 \times 512$), pool4, fc1
($131072 \times 20$).
\clutrender (Sect.~\ref{sec:exp_diagnostic}):
conv1-conv3 ($128 \times 128 \times 64$), pool1, conv4-conv6 ($64
\times 64 \times 128$), pool2, conv7-conv9 ($32 \times 32 \times
256$), pool3, conv10-conv12 ($16 \times 16 \times 512$), pool4, fc1
($32768 \times 14$).
MNIST (Sect.~\ref{sec:exp_diagnostic}): conv1-conv3
($120 \times 120 \times 32$), pool1, conv3-conv4 ($60 \times 60 \times
64$), pool2, conv5-conv6 ($30 \times 30 \times 128$), pool3, fc1
($28800 \times 10$).}
%

The discriminative loss
functions $L_{m_{k}}$ and $L_{u}$ are instantiated as softmax-cross
entropy or element-wise sigmoid-cross entropy for joint or independent class
prediction, respectively.
Since $\mathcal{L}_c$ 
is of a standard form, we can optimize it like any CNN via SGD
(specifically, using the ADAM optimizer~\cite{ADAM} and
Tensorflow~\cite{tensorflow2015-whitepaper}).

Empirically, we find that best performance is achieved
when applying $\mathcal{L}_c$ to the topmost
convolutional and pooling layers of the network (i.e., $\lambda_{n}$
is zero on most early layers).
We believe this to be an artifact of the CNN needing a certain minimum
number of layers and corresponding representational power to
successfully discriminate between relevant 
and irrelevant (background) pixels.

In practice, we create only two weight-sharing CNNs
(independent of the number of object training instances): one which
sees $1$ randomly selected out of $K$ objects in the input
image, and one which sees the entire scene. Empirically, this model is
only about $50\%$ slower to train than a standard CNN. The parameter
space is just that of a single CNN due to weight sharing.
$\gamma$ is fixed to $.5$. 

\section{Experiments}
\label{sec:experiments}
%
%
%
%
%
In this section, we give a detailed experimental evaluation of our
approach for teaching compositionality to CNNs, highlighting its
ability to improve performance over standard CNN training on both
synthetic (Sect.~\ref{sec:exp_diagnostic}) and real images
(\mscoco~\cite{lin14eccv}, Sect.~\ref{sec:exp_mscoco}).
Our emphasis lies on providing an in-depth analysis of the
contributions of different components of our compositional objective
as well as quantifying the impact of object context on performance.
%

\subsection{Datasets and metrics}
\label{sec:exp_datasets}
%
\paragraph{Rendered 3D objects.}
We perform diagnostic experiments on two novel datasets of rendered 3D
objects. We use rendered datasets so we have maximum control over 
the statistics of our image data in terms of depicted objects and context
(notably, segmentation masks come for free in this setting).
Specifically, the datasets are based on $12$ 3D object
classes (e.g., car, bus, boat, or airplane), with $\approx 20$
object instances per category, each rendered from $\approx 50$
different viewpoints (uniform sampling of the upper viewing
half-sphere) in front of $20$ different real-image backgrounds.

The first dataset, termed \clutrender, has $1,600$ images depicting
single objects in front of random backgrounds. The second dataset, termed
\multirender, has $800$ images of multiple objects with varying
degrees of occlusion (see Fig.~\ref{fig:VGGNotComp}).
For both datasets, we distinguish between a category-level recognition
setting and easier variants (\clutrenderinst, \multirenderinst) that
allow the set of 3D object instances seen during training and test to
be non-empty (whereas the set of views of the same instance has to be
empty). In both cases, we ensure that the backgrounds seen in
training ($80\%$ of the images) and test ($20\%$) are distinct.
%
%
\paragraph{MNIST.}
We create two variants of the popular MNIST
dataset~\cite{MNIST}, in analogy to the two aforementioned 3D object
datasets. The first variant, \clutmnist, depicts
individual MNIST characters in front of randomized, cluttered
backgrounds (we use the standard train/test split). The second
variant, \multimnist, depicts multiple characters with varying degrees
of overlap, against these backgrounds.
\paragraph{\mscocosub.}
\mscoco~\cite{lin14eccv} constitutes
a move away from ``iconic'' views of objects towards a
dataset in which objects frequently occur in their respective natural
contexts.
For most experiments, we focus on subsets 
of \mscoco training and validation (for testing) images
that contain at least one of
$20$ diverse object classes\footnote{The first 20 classes in the
original MS-COCO ordering w/o {\em person}.}
(see Fig.~\ref{fig:mscoco_perclass}) and
further restrict the set of images to ones with sufficiently large
object instances of at least $7,000$ pixels. This
results in $22,476$ training and $12,245$ test images.
%

In addition, we quantify the impact of context on classification
performance by defining two further test sets. The first
test set is the full validation set of \mscoco. Here, we measure
classification performance on object instances of different
sizes (small, medium, large) as defined for the \mscoco detection
challenge\footnote{\url{http://mscoco.org/dataset/\#detections-eval}}.
%
%
In order to make the performance comparable, 
we stratify the number of positive and negative examples by randomly
sampling 20 negatives for each positive example.
%
%
%

The second test set examines object instances {\em in} and {\em
out of context} (see Fig.~\ref{fig:qualitative}~(b)).
We start with all test images 
from \mscocosub. For each object instance $o$ of a category $c$
occurring in that set, we create two positive examples, one by
cropping $o$ and placing it in front of a new random test image that
does not have $c$ in it (this will be the out-of-context set), and one
by leaving $o$ in its original context (the in-context set). For both,
we add as negative examples all images where $c$ does not occur.
%
%
\paragraph{Metrics.}
All experiments consider image-level classification tasks, not object
class detection or localization.
For diagnostic experiments (Sect.~\ref{sec:exp_diagnostic}), we
evaluate performance as the average fraction of
correctly predicted object classes among the top-$k$ scoring network
predictions, where $k$ is the number of objects in a given image. 
For \mscocosub (Sect.~\ref{sec:exp_mscoco}), we treat object classes
separately and report (mean) average precision (AP) over independent
binary classification problems.
In all cases, we monitor performance on a held-out test set over
different epochs as training progresses, and report both the
resulting curves and best achieved values per method
(Fig.~\ref{fig:convergence}, Fig~\ref{fig:mscoco}).
\subsection{Methods}
\label{sec:exp_methods}
In this section, we evaluate the following
baselines and variations on our compositional training
technique (see Sect.~\ref{sec:training}). For the sake of clean
comparison, we always train all networks from scratch (i.e., we do not
use pre-training of any form).\\
\textbf{\compfull.} Our main architecture, where $m'_{k}$ is chosen to
be equal to a block of all $1$s but with the locations of objects
other than the $k$\textit{th} object set to $0$s.\\
\textbf{\compobjonly.} Like \compfull, but with $m'_{k}$ equal to $m_{k}$
(this penalizes any shifts in activations within the object
region but does not discourage background activations).\\
\textbf{\compnomask.} Like \compfull, except that the masked CNNs do not
apply $m_{k}$ to any of their activations.\\
\textbf{\baseline.} Architecture with the same layer sizes as
\compfull but without compositional objective terms -- a ``standard" CNN. \\ 
\textbf{\baselineaug.} Like \baseline, except 
for each batch we make half of the images be a
single object shown in isolation against a black background
and the other half be the 
raw images of the same objects in the same locations against
cluttered background.
This method has access to the same 
information as \compfull (it knows about the object mask), but without
any compositional objective. \\
%
\textbf{\baselinereg.} Like \baseline, but with
dropout~\cite{srivastava14jmlr} and $l_2$-regularization. \\
\textbf{\baselineaugreg.} Like \baselineaug, but with dropout
and $l_2$-regularization.
%
%
%

\subsection{Diagnostic experiments on synthetic data}
\label{sec:exp_diagnostic}
\begin{figure*}[t]
  \centering
  \begin{subfigure}[b]{.245\textwidth}
  \centering
    \includegraphics[width=\textwidth]{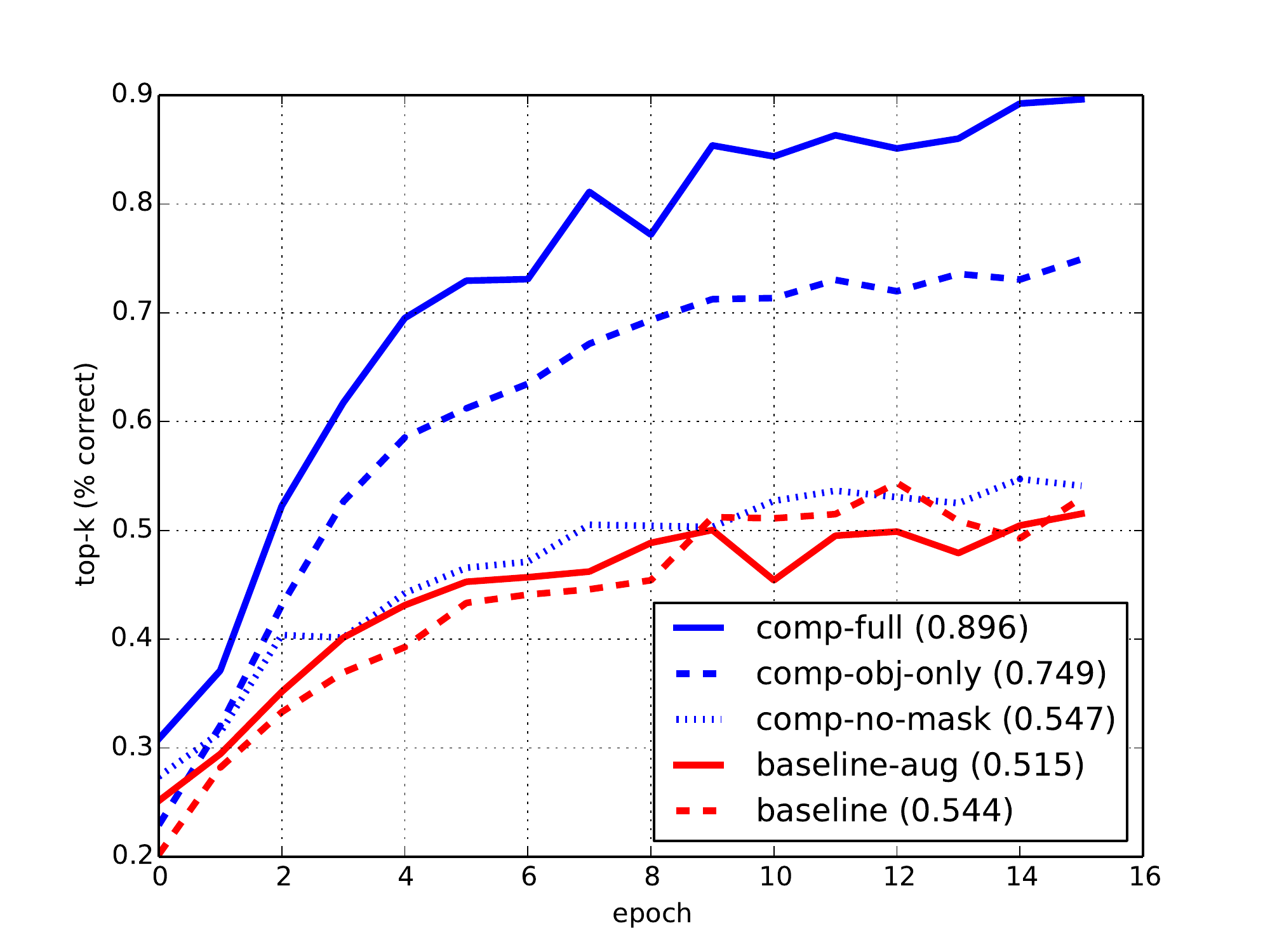}
    \caption{\clutrenderinst}
    \label{fig:conv_clutrenderinst}
  \end{subfigure}
  \begin{subfigure}[b]{.245\textwidth}
  \centering
    \includegraphics[width=\textwidth]{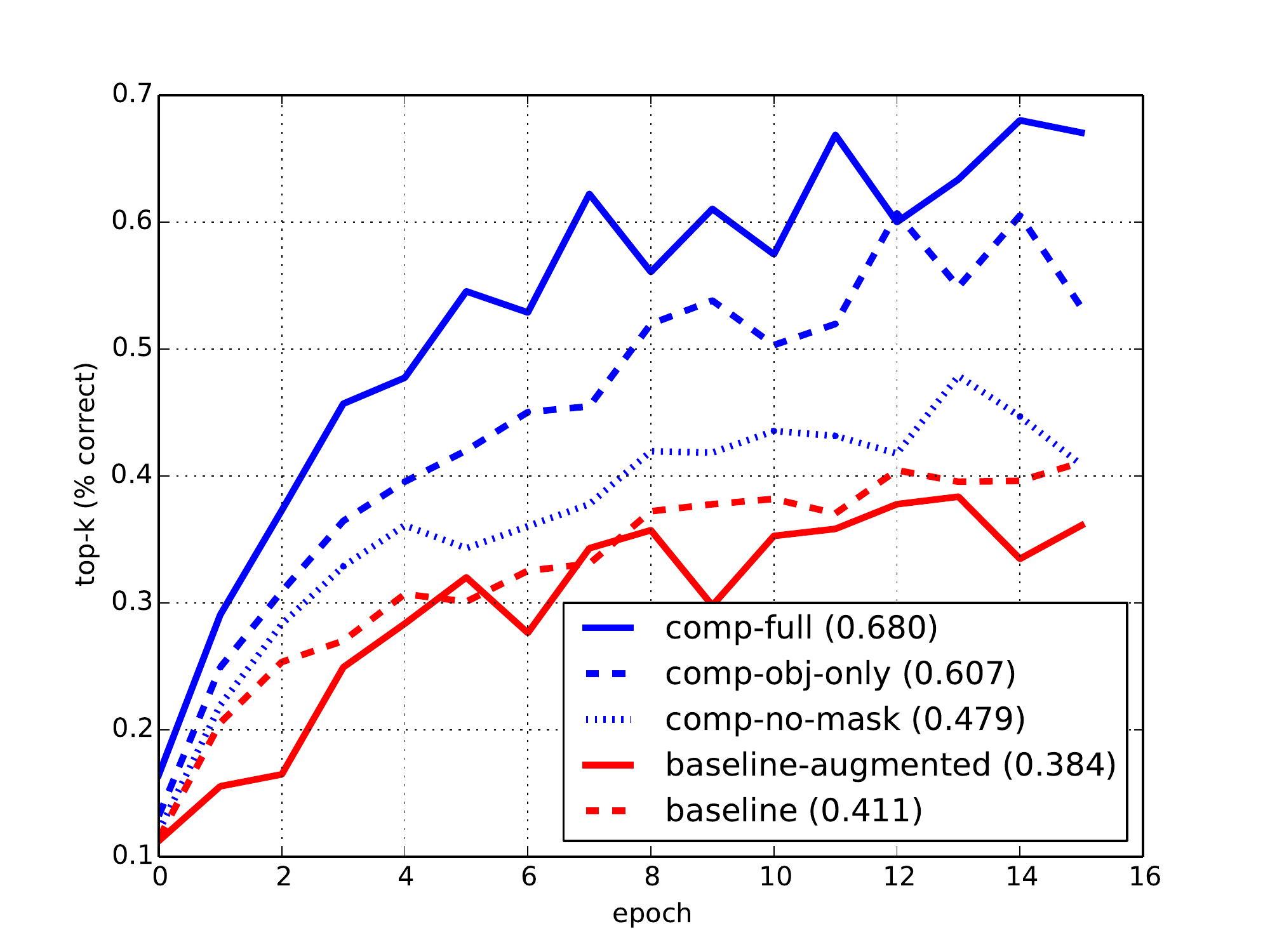}
    \caption{\clutrender}
    \label{fig:conv_clutrender}
  \end{subfigure}
  \begin{subfigure}[b]{.245\textwidth}
  \centering
    \includegraphics[width=\textwidth]{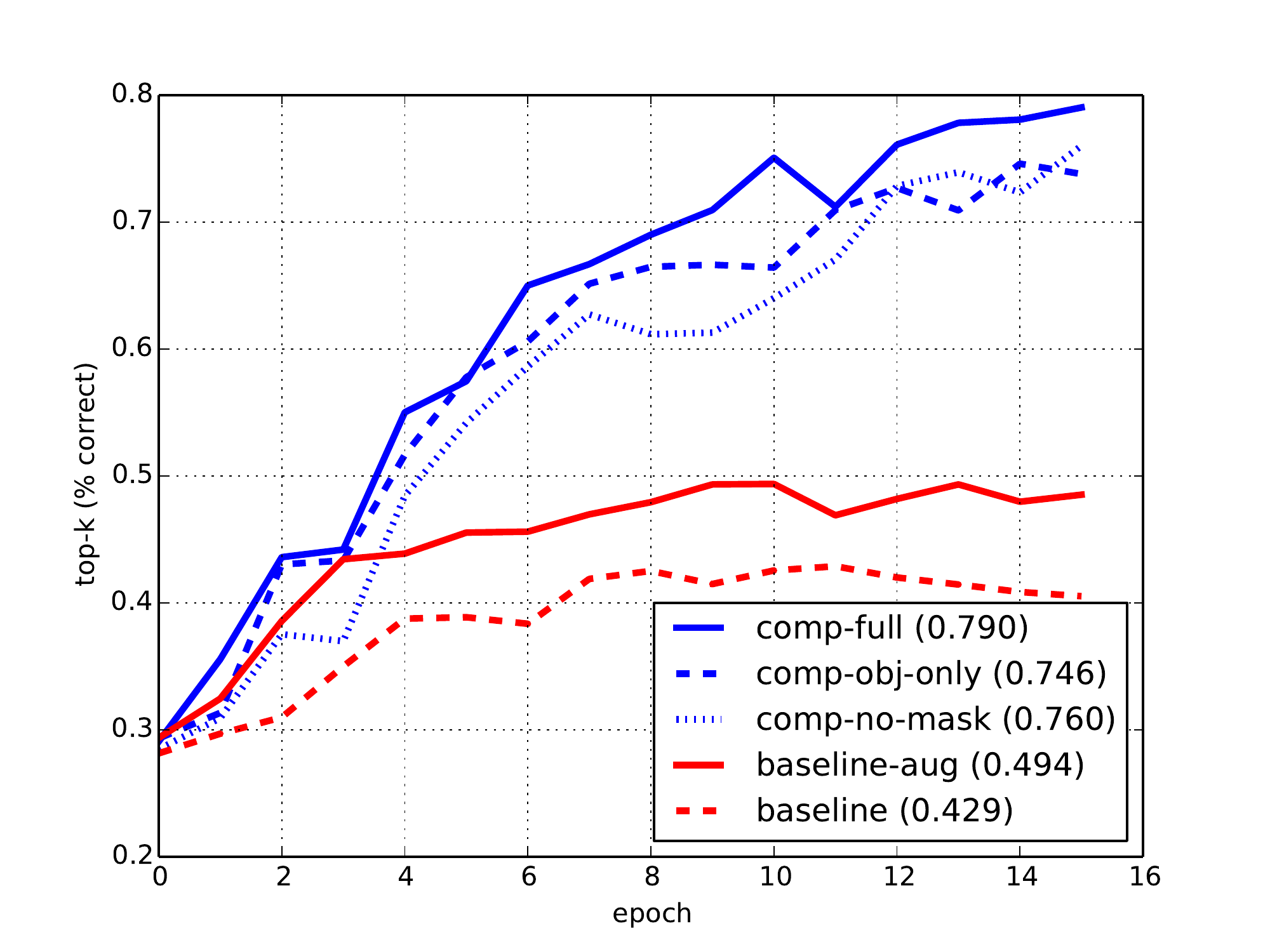}
    \caption{\multirenderinst}
    \label{fig:conv_multirenderinst}
  \end{subfigure}
  \begin{subfigure}[b]{.245\textwidth}
  \centering
    \includegraphics[width=\textwidth]{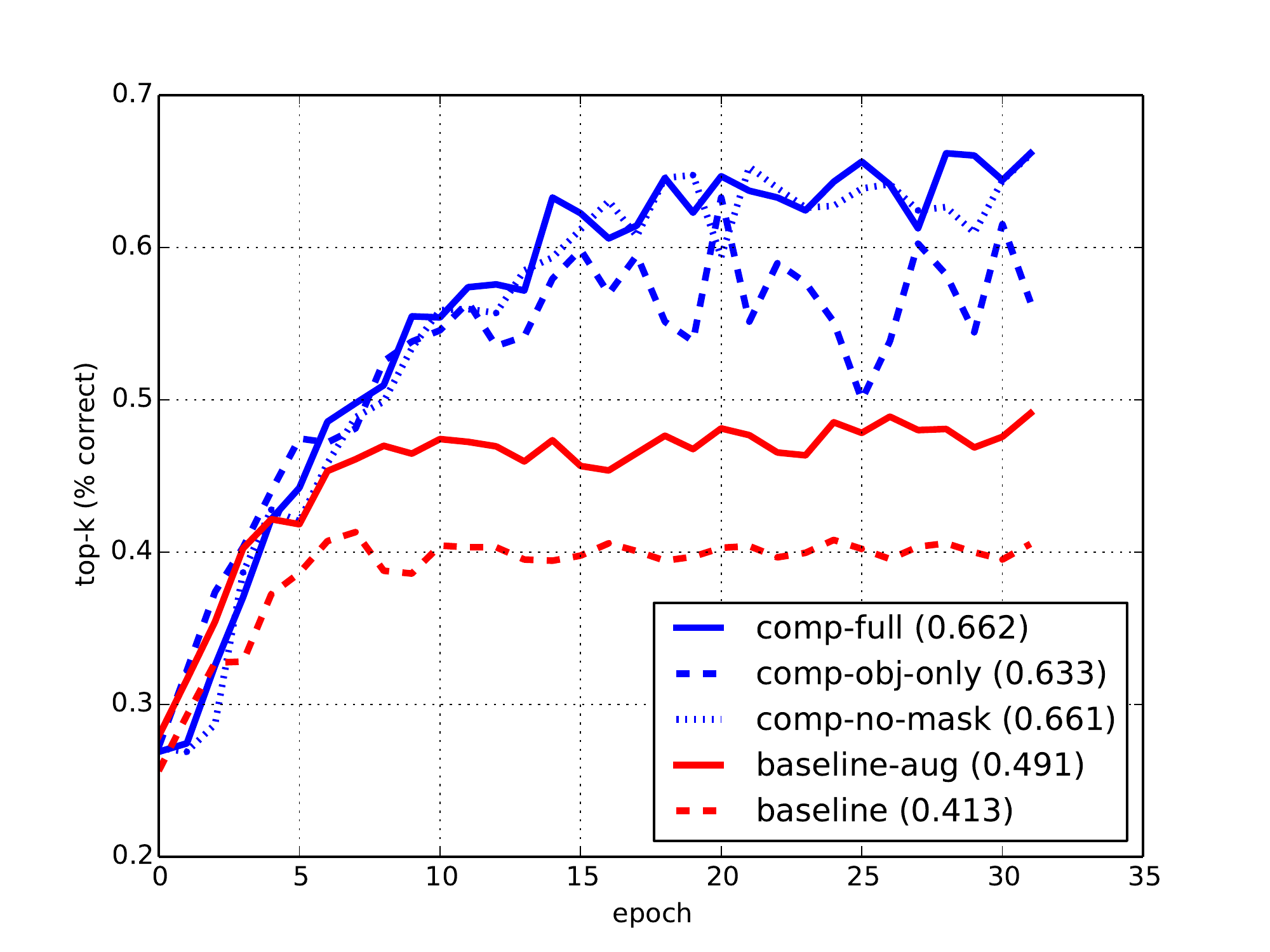}
    \caption{\multirender}
    \label{fig:conv_multirender}
  \end{subfigure} \\
  \begin{subfigure}[b]{.245\textwidth}
  \centering
    \includegraphics[width=\textwidth]{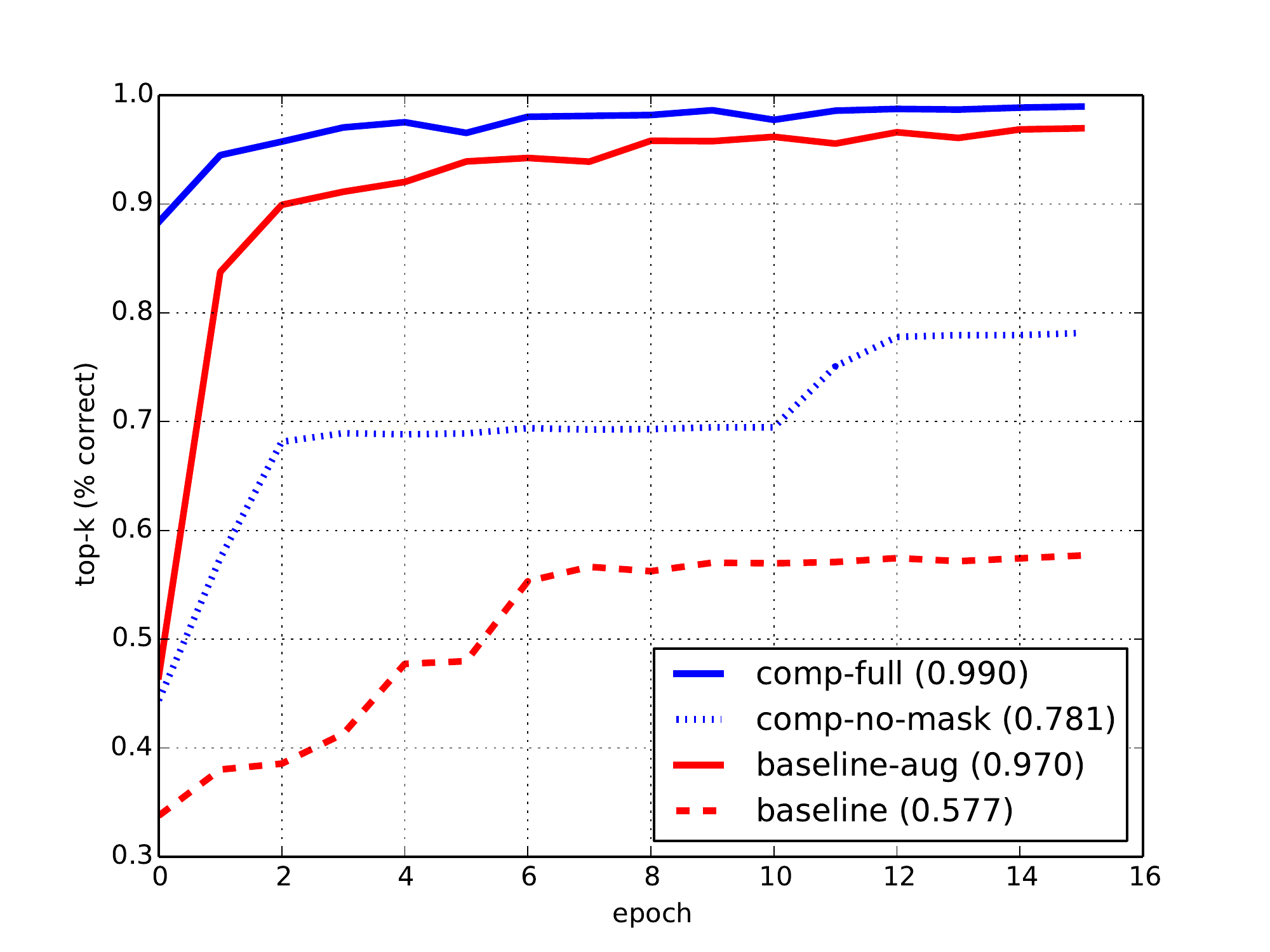}
    \caption{\clutmnist}
    \label{fig:conv_clutmnist}
  \end{subfigure}
  \begin{subfigure}[b]{.245\textwidth}
  \centering
    \includegraphics[width=\textwidth]{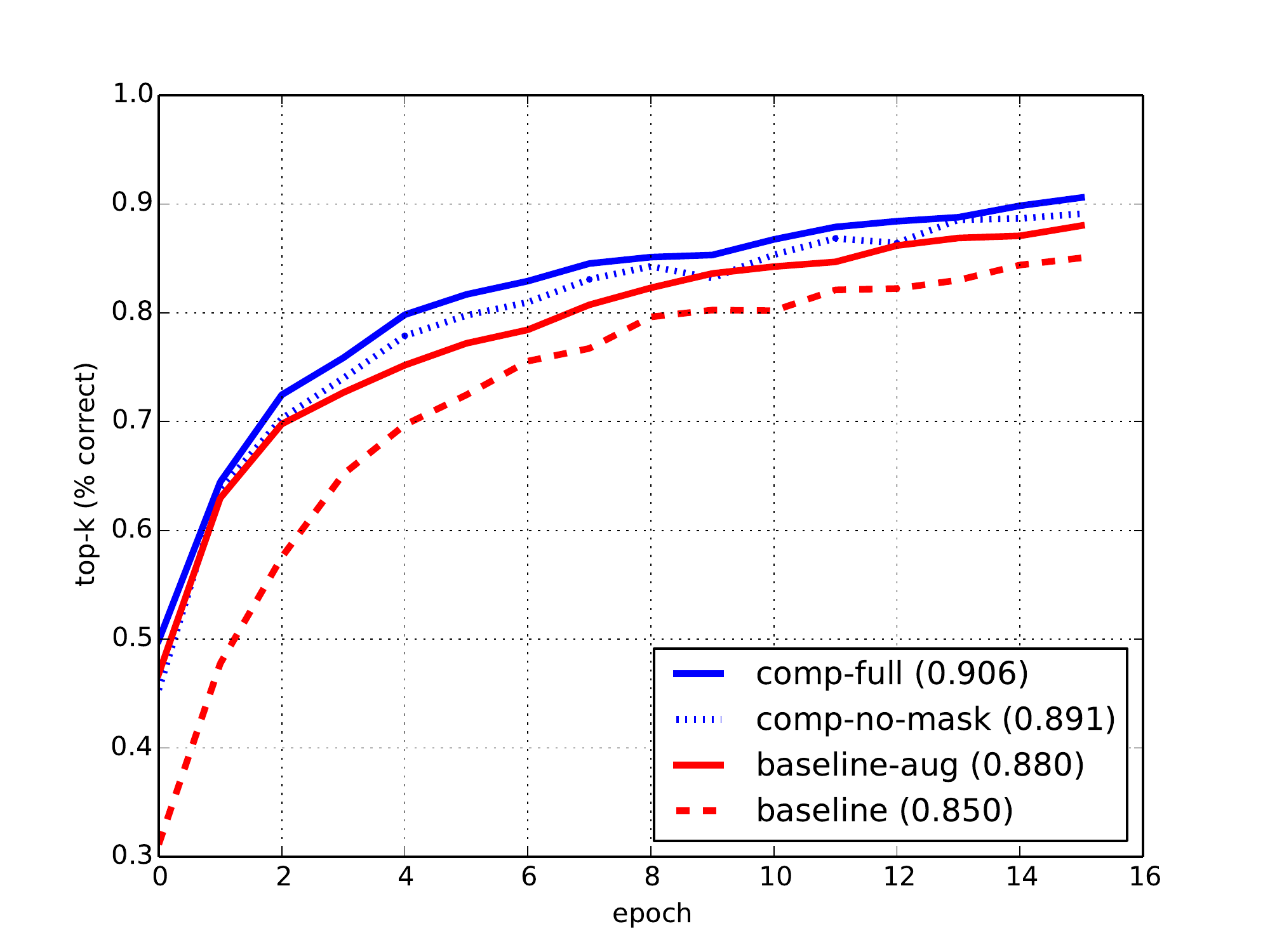}
    \caption{\multimnist}
    \label{fig:conv_multimnist}
  \end{subfigure}
   \begin{subfigure}[b]{.23\textwidth}
   \centering
     \includegraphics[width=\textwidth]{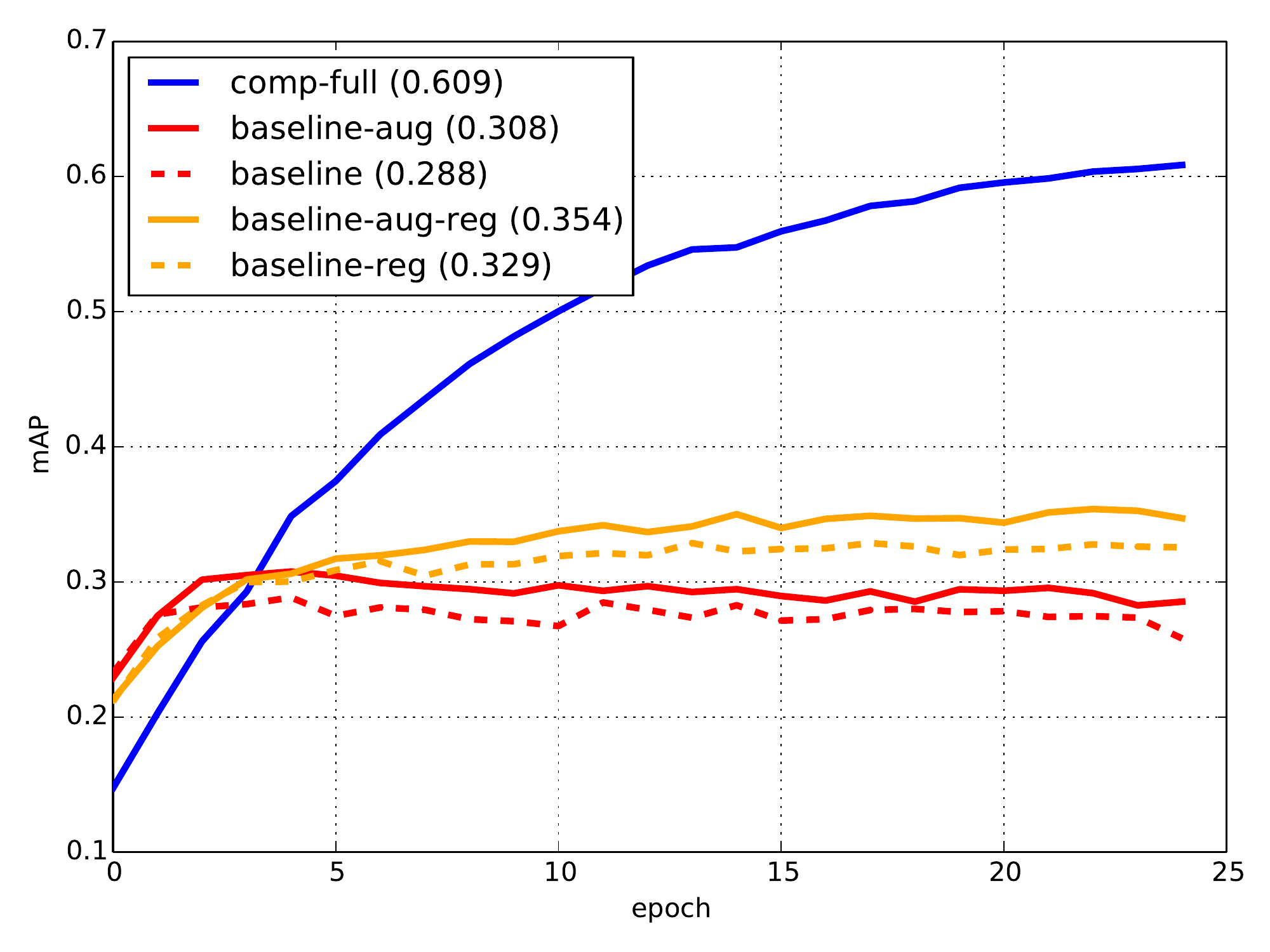}
     \caption{\mscocosub}
     \label{fig:conv_mscoco}
   \end{subfigure}
   \begin{subfigure}[b]{.23\textwidth}
   \centering
   \begin{scriptsize}
   \setlength{\tabcolsep}{2pt}
   \begin{tabular}{c|c}
   \hline
   {\em method} & {\em localization}  \\
                & {\em accuracy}      \\
   \hline
   \hline
   \compfull              & {\bf 0.368} \\
   \hline
   \baselineaug           &  0.256 \\
   \hline
   VGG~\cite{Zisserman15} & 0.330 \\
   \hline
   \end{tabular}
   \vspace{.7cm}
   \end{scriptsize}
   %
     \caption{\mscocosub}
     \label{tab:compositionality}
   \end{subfigure}
  %
  %
\caption{Test performance on rendered 3D objects (a-d),
  MNIST (e-f), and \mscocosub (g), as training progresses over epochs
  (best performance per curve given in parentheses; see
  Sect.~\ref{sec:exp_diagnostic}
  and~\ref{sec:exp_mscoco}). Localization accuracy (h)
  (Sect.~\ref{sec:exp_mscoco}).}
\label{fig:convergence}
\end{figure*}
We commence by comparing the performance of different variants of our
compositional objective and the corresponding baselines
(Sect.~\ref{sec:exp_methods}) in a diagnostic setting on synthetic data.
In order to assess both best case performance and convergence
behavior, we plot test performance vs. training
epochs in Fig.~\ref{fig:conv_clutrenderinst}
through~\ref{fig:conv_multimnist}. The respective best performance per
curve is given in parentheses in plot legends.
Fig.~\ref{fig:qualitative} and~\ref{fig:guided_backprop} give
qualitative results.
\paragraph{Rendered 3D objects.}
In Fig.~\ref{fig:convergence}, we observe that all variants of
compositional CNNs (blue curves)
perform consistently better than the baselines (red curves), both
per epoch and in terms of best case performance.

Our full model,~\compfull, performs overall best (blue-solid).
It outperforms the best baseline by between $17.1\%$ (\multirender,
Fig.~\ref{fig:conv_multirender}) and $35.2\%$
(\clutrenderinst, Fig.~\ref{fig:conv_clutrenderinst}).
Performance drops for \compobjonly (blue-dashed) by $14.7\%$
(\clutrenderinst, Fig.~\ref{fig:conv_clutrenderinst}), $7.3\%$
(\clutrender, Fig.~\ref{fig:conv_clutrender}), $4.4\%$ (\multirenderinst,
Fig.~\ref{fig:conv_multirenderinst}), and $2.9\%$ (\multirender,
Fig.~\ref{fig:conv_multirender}), respectively.
%
%
\compnomask (blue-dotted) performs worst among our models, but still
outperforms the best baseline by $0.3\%$ (\clutrenderinst,
Fig.~\ref{fig:conv_clutrenderinst}), $6.8\%$ (\clutrender,
Fig.~\ref{fig:conv_clutrender}), $26.6\%$ (\multirenderinst,
Fig.~\ref{fig:conv_multirenderinst}), and $17.0\%$ (\multirender,
Fig.~\ref{fig:conv_multirender}), respectively.

As expected, the baseline benefits from observing additional masked
training data mostly for images with multiple objects:
\baseline (red-dashed) and \baselineaug (red-solid) perform
comparably on \clutrenderinst and \clutrender, but \baselineaug
improves over \baseline on \multirenderinst (by $6.2\%$)
and \multirender (by $7.8\%$).
In terms of convergence, the compositional CNNs
(blue curves) tend to stabilize later than the baselines (red
curves).
%
%
%

\paragraph{MNIST.} In Fig.~\ref{fig:conv_clutmnist} and
~\ref{fig:conv_multimnist}, the absolute
performance differences between our compositional CNNs and the
corresponding baselines are less clear cut, but still highlight the
importance of the compositional objective when object masks are used
(\compfull outperforms \baselineaug by $2.0\%$ on \clutmnist and by
$2.6\%$ on \multimnist). Without reapplying the masks to the
activations, performance decreases, but the trend
remains (\compnomask is better
than \baseline by $20.4\%$ and $4.1\%$, respectively).


\subsection{Experiments on real-world data (\mscoco)}
\label{sec:exp_mscoco}
\begin{figure*}[t]
  \captionsetup[subfigure]{justification=centering}
  \centering
  \begin{subfigure}[b]{.65\textwidth}
    \centering
    \includegraphics[width=\textwidth]{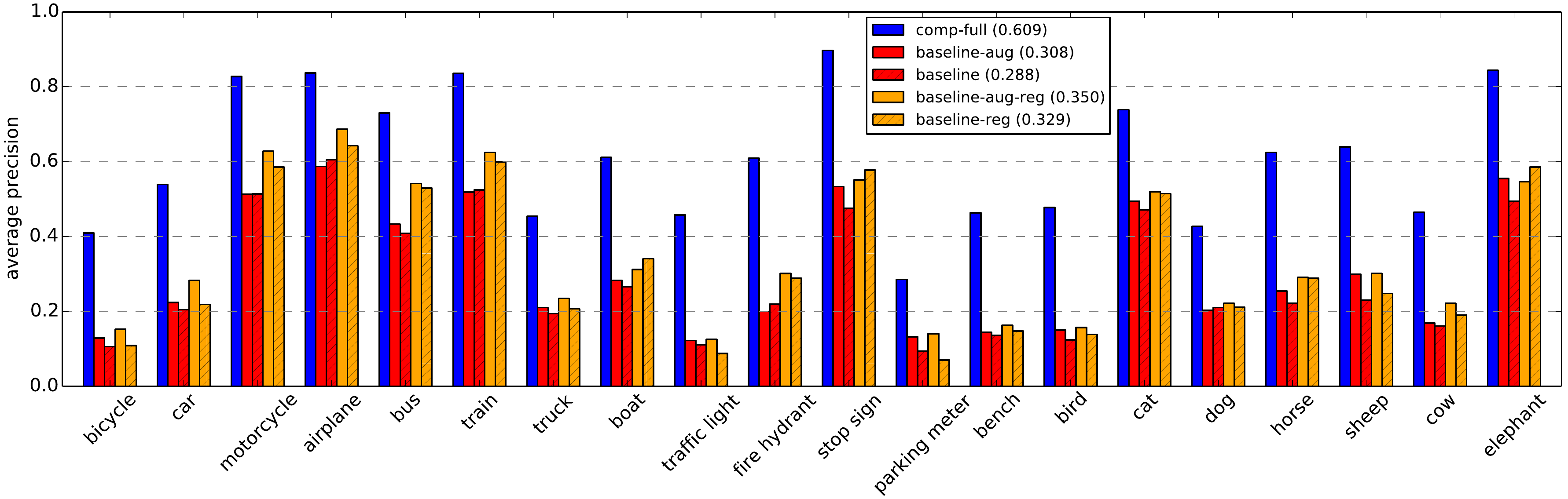}
    \caption{\mscocosub test performance (AP) per object category
  (corresponding to the last epoch for \compfull~/~best case performance
  for baselines reported in Fig.~\ref{fig:conv_mscoco}).}
    \label{fig:mscoco_perclass}
  \end{subfigure}
  \begin{subfigure}[b]{.24\textwidth}
    \centering
    \includegraphics[width=\textwidth]{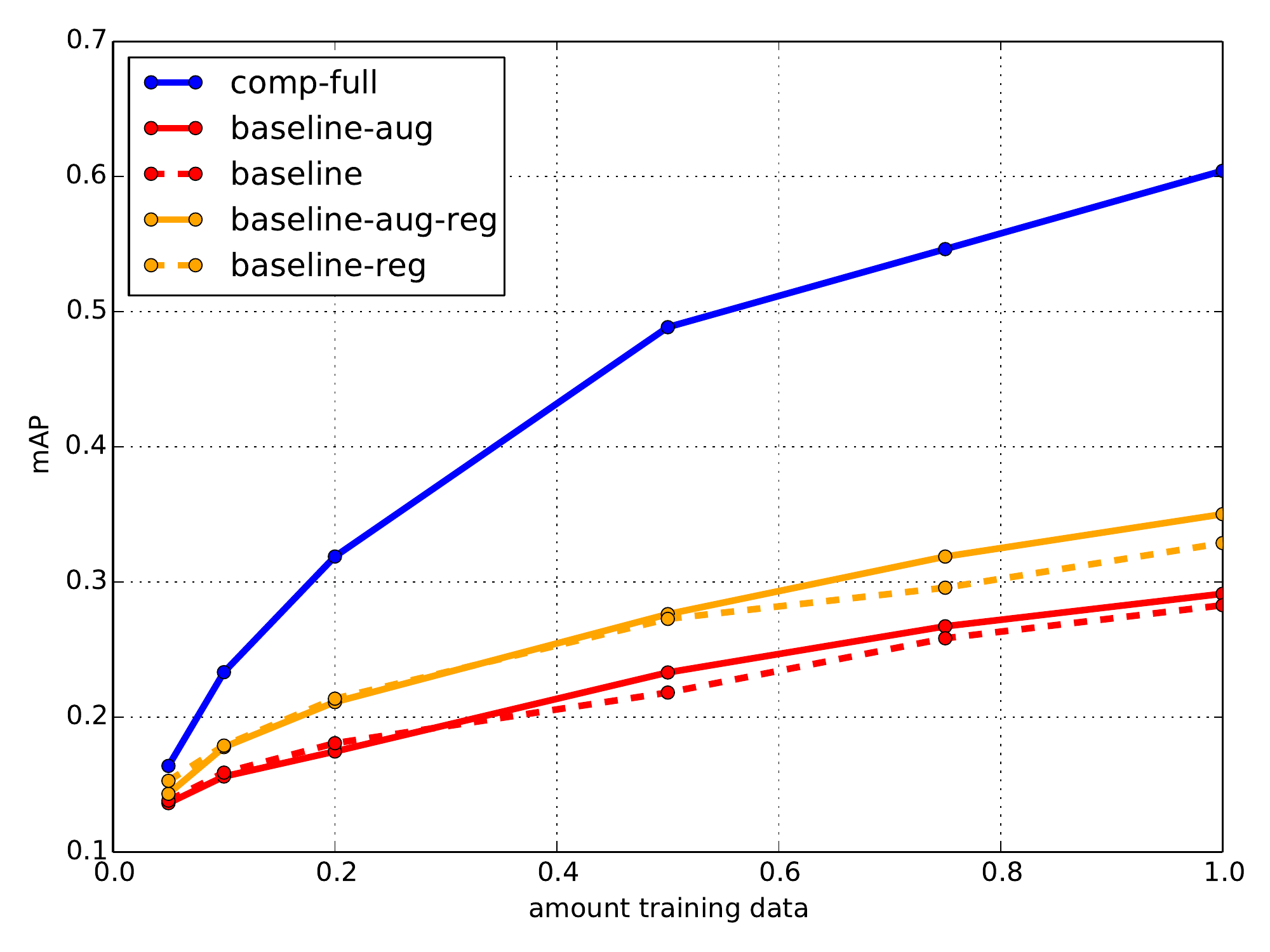}
    \vspace{.04cm}
    \caption{Performance (mAP) for fractions of training data.}
    \label{fig:mscoco_traininig}
  \end{subfigure} \\
  \begin{subfigure}[b]{.32\textwidth}
    \centering
    \includegraphics[width=\textwidth]{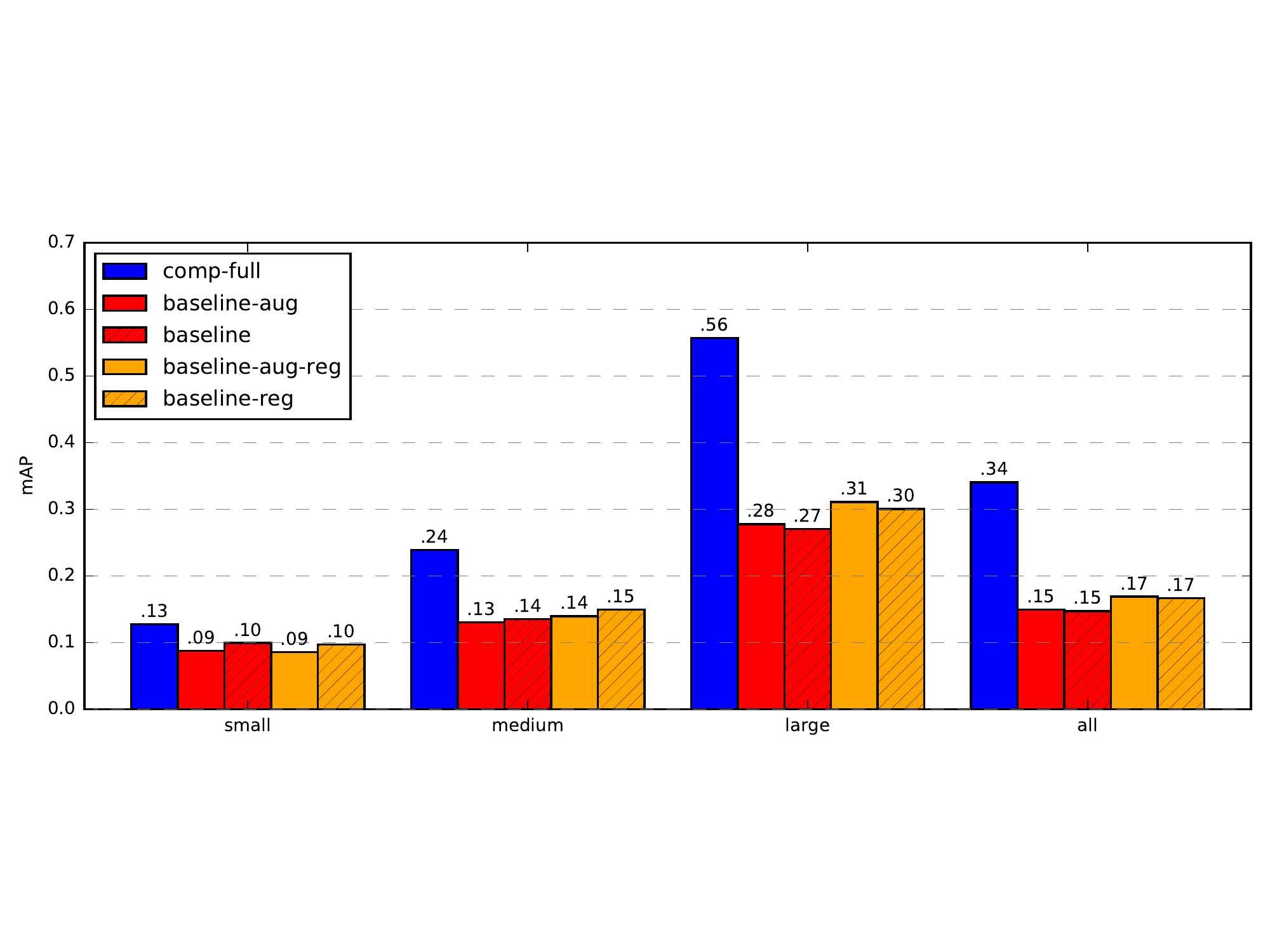}
    \caption{\mscocosub performance for objects of different sizes
  (small, medium, large).}
    \label{fig:mscoco_sizes}
  \end{subfigure}
  \begin{subfigure}[b]{.64\textwidth}
    \centering
    \includegraphics[width=\textwidth]{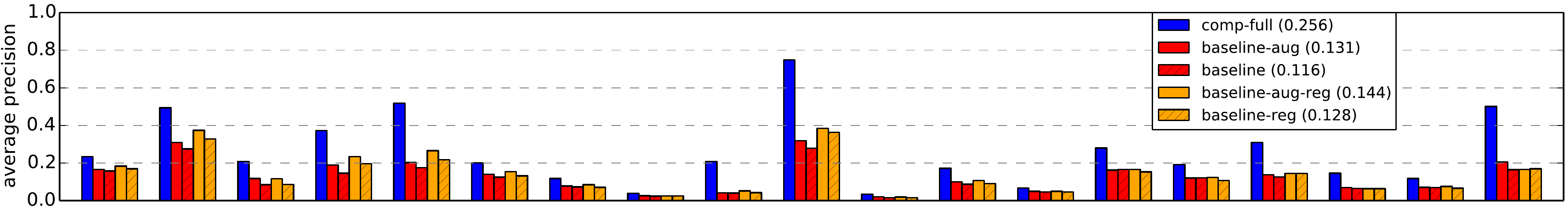} \\
    \includegraphics[width=\textwidth]{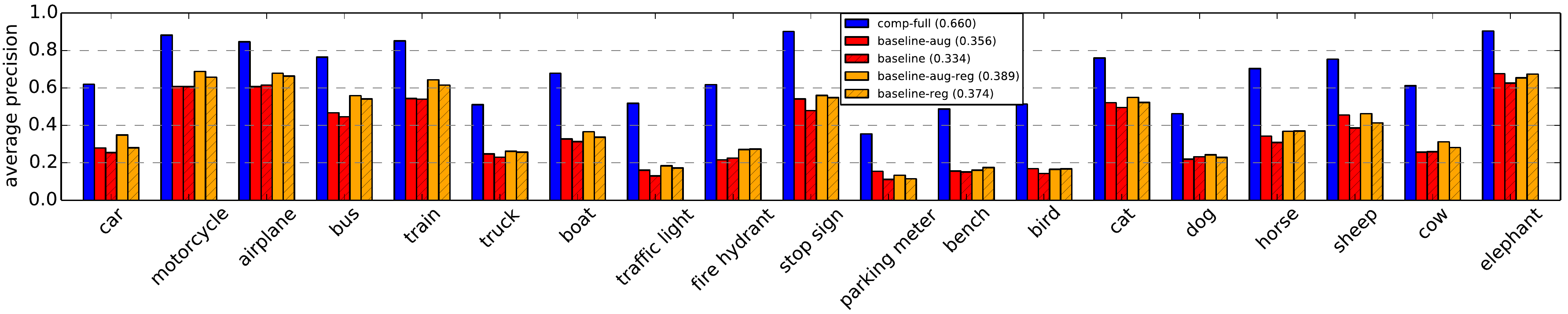}
    \caption{\mscocosub performance for objects in-context (bottom)
    and \newline out-of-context (top).
    See Sect.~\ref{sec:exp_mscoco} for details.}
    \label{fig:mscoco_context}
  \end{subfigure}
  \caption{\mscocosub performance per object class (a), training set
  size (b), object instance size (c), and context (d).}
  \label{fig:mscoco}
\end{figure*}
We proceed to evaluating our best performing method ~\compfull on the
real-world images of \mscoco (Sect.~\ref{sec:exp_datasets}).
%
We compare to the same baselines as before, plus two baselines with
dropout~\cite{srivastava14jmlr} and $l_2$-regularization (see
Sect.~\ref{sec:exp_methods}).
Specifically, we report performance for \compfull at convergence (last
epoch, see Fig.~\ref{fig:conv_mscoco} for convergence behavior); for
all baselines we consider the best performing model over all epochs.
%
%
Fig.~\ref{fig:mscoco} gives details
w.r.t. individual object categories (\ref{fig:mscoco_perclass}),
amount of training data (\ref{fig:mscoco_traininig}), size of
object instances (\ref{fig:mscoco_sizes}), and context
(\ref{fig:mscoco_context}).
\paragraph{\mscocosub.}
In Fig.~\ref{fig:conv_mscoco}, \compfull (blue-solid) outperforms the
best baseline~\baselineaugreg
(orange-solid) by a significant margin of $25.5\%$, confirming the benefit of
the compositionality objective in a real-world setting. 
The added regularization improves the performance of
the baselines only moderately, by $4.6\%$ (\baselineaug) and $4.1\%$
(\baseline), respectively.
In Fig.~\ref{fig:mscoco_perclass}, we see that \compfull performs
better than the baselines for every single category, improving
performance by up to $32\%$ (for {\em stop sign}).
%
%
\begin{table}[ht!]
\centering
\begin{scriptsize}
\begin{tabular}{c|c|c|c}
\hline
{\em method} & {\em in-context} & {\em out-of-context} & {\em ratio} \\
\hline
\hline
\compfull       & {\bf 0.660}  & {\bf 0.256} & {\bf 0.39} \\
\hline
\baselineaug    & 0.356 & 0.131 & 0.37 \\
\baseline       & 0.334 & 0.116 & 0.35 \\
\hline
\baselineaugreg & 0.389 & 0.144 & 0.37 \\
\baselinereg    & 0.374 & 0.128 & 0.34 \\
\hline
\end{tabular}
\end{scriptsize}
\caption{Relative performance ratio on \mscocosub.}
\label{tab:rel_performance}
\end{table}
%
Fig.~\ref{fig:mscoco_traininig} gives results for varying amounts of
training data ($5,10,20,50,75,100\%$). 
\compfull (blue-solid) clearly outperforms the baselines (orange and red curves)
for all plotted amounts, with an increasing performance gap as
training data increases.
%
\paragraph{Object sizes and context.}
Fig.~\ref{fig:mscoco_sizes} gives the performance when testing the
respective models trained on the training portion of \mscocosub on all
images of \mscoco and evaluating them on object instances of
different sizes (small, medium, large, all; see
Sect.~\ref{sec:exp_datasets}).

We observe that the compositional objective improves performance over
the baselines consistently for all sizes. The improvement is most
pronounced for large object instances ($25\%$, \compfull
vs. \baselineaugreg), decreases for medium ($9\%$, \baselinereg),
and almost vanishes for small objects ($3\%$, \baselinereg).
This ordering is in line with the intuition that the compositional
objective encourages activations to be context invariant: as context
becomes more important with decreasing object size, the advantage of
context invariance decreases.

\begin{figure*}[t]
  \centering
  \scriptsize
  \setlength{\tabcolsep}{1.1pt}
  \renewcommand{\arraystretch}{1}
  \newcolumntype{C}{>{\centering\arraybackslash} m{.08\textwidth}}
  \begin{tabular}{ccc}
  \begin{tabular}{m{.14\textwidth}CC|CCC|CCC}
    %
    %
    \multirender (bus, gun) &
    \includegraphics[width=.08\textwidth]{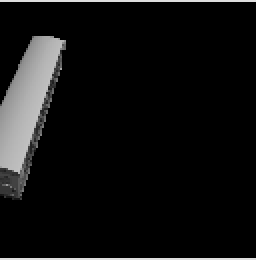} &
    \includegraphics[width=.08\textwidth]{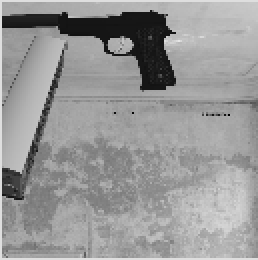} & 
    \includegraphics[width=.08\textwidth]{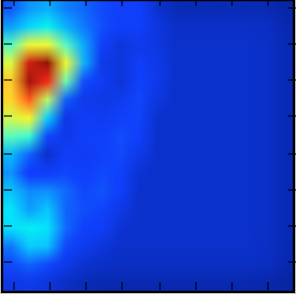} &
    \includegraphics[width=.08\textwidth]{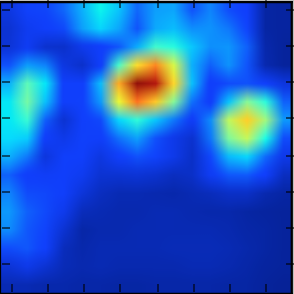} &
    \includegraphics[width=.08\textwidth]{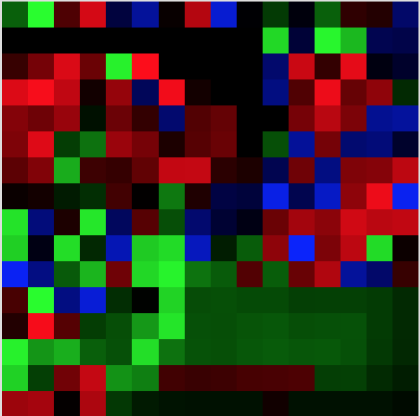} &
    \includegraphics[width=.08\textwidth]{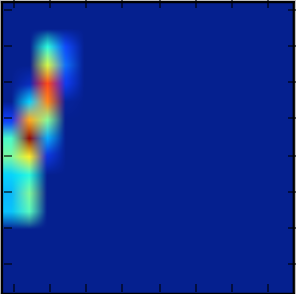} &
    \includegraphics[width=.08\textwidth]{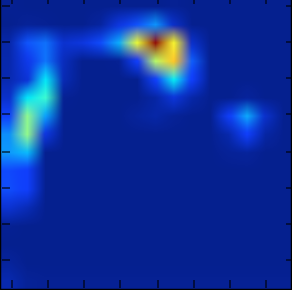} &
    \includegraphics[width=.08\textwidth]{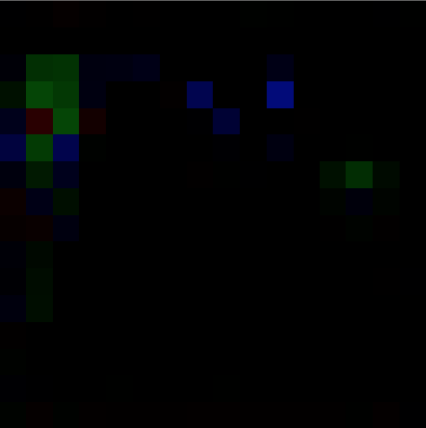} \\
    %
    %
    \multimnist (8, 5, 3) &
    \includegraphics[width=.08\textwidth]{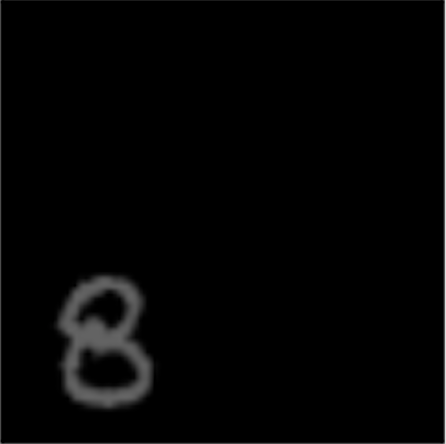} &
    \includegraphics[width=.08\textwidth]{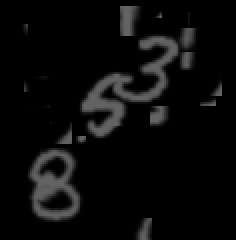} &
    \includegraphics[width=.08\textwidth]{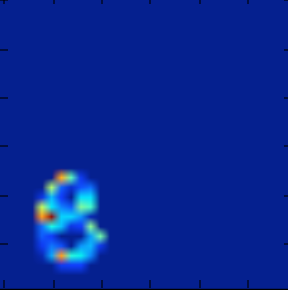} &
    \includegraphics[width=.08\textwidth]{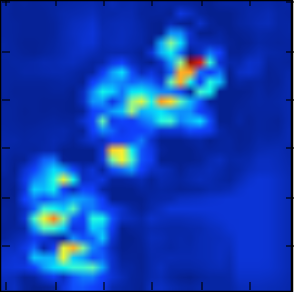} &
    \includegraphics[width=.08\textwidth]{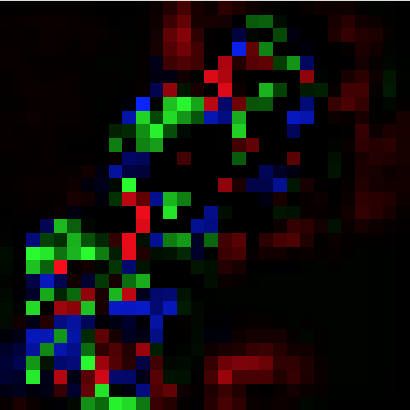} &
    \includegraphics[width=.08\textwidth]{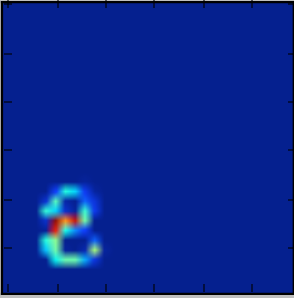} &
    \includegraphics[width=.08\textwidth]{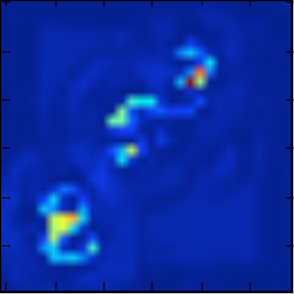} &
    \includegraphics[width=.08\textwidth]{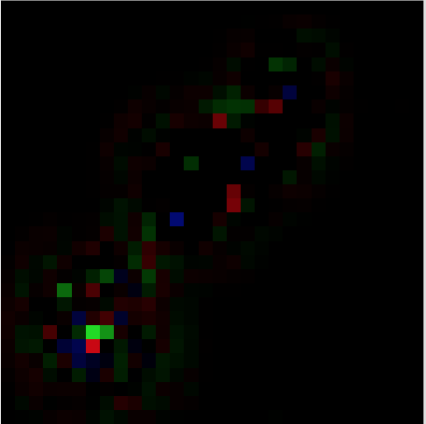} \\
    \mscocosub (horse) & 
    \includegraphics[width=.08\textwidth]{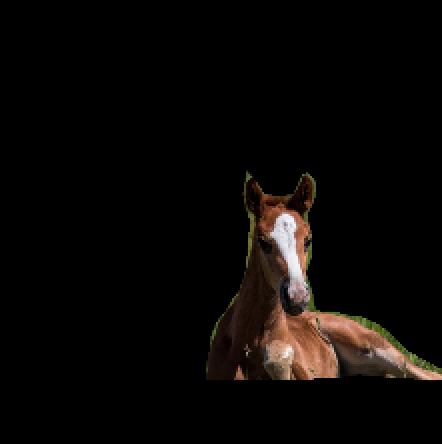} &
    \includegraphics[width=.08\textwidth]{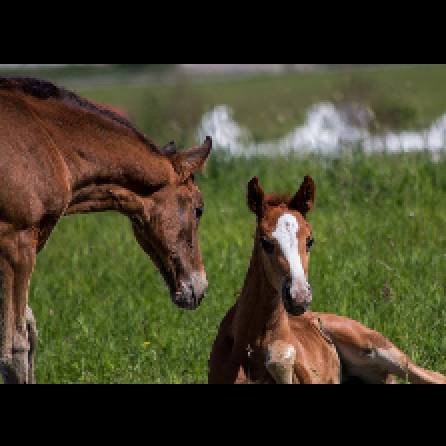} &
    \includegraphics[width=.08\textwidth]{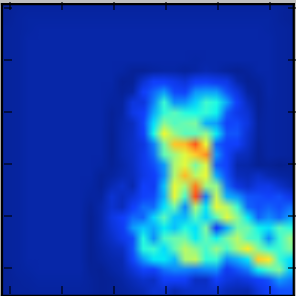} &
    \includegraphics[width=.08\textwidth]{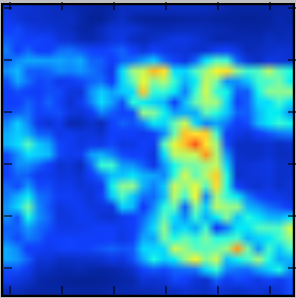} &
    \includegraphics[width=.08\textwidth]{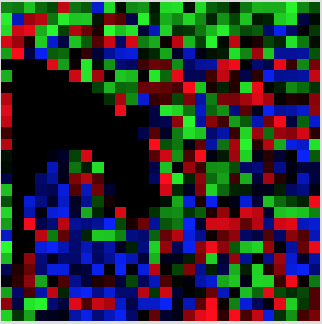} &
    \includegraphics[width=.08\textwidth]{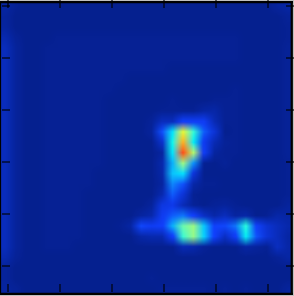} &
    \includegraphics[width=.08\textwidth]{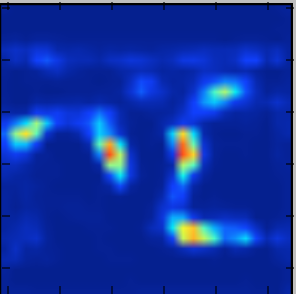} &
    \includegraphics[width=.08\textwidth]{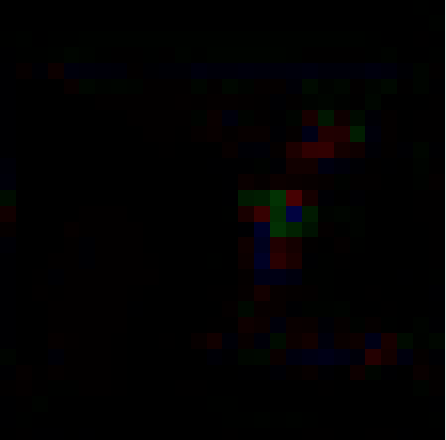} \\
    %
    %
(a) & masked & unmasked & act. masked & unmasked & act. shift & act. masked
    & unmasked & act. shift \\
    & input & input & \multicolumn{3}{c|}{\baselineaug} & \multicolumn{3}{c}{{\bf \compfull (ours)}}
  \end{tabular} & {~~~~~~} &
  \begin{tabular}{cc}
    \includegraphics[width=.06\textwidth]{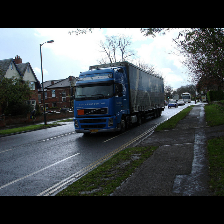} &
    \includegraphics[width=.06\textwidth]{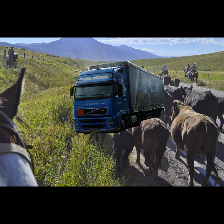} \\
    \includegraphics[width=.06\textwidth]{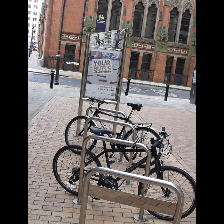} &
    \includegraphics[width=.06\textwidth]{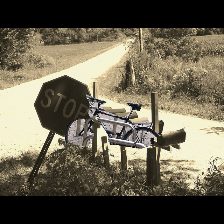} \\
    \includegraphics[width=.06\textwidth]{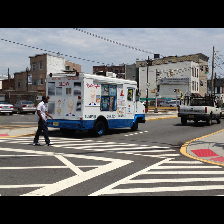} &
    \includegraphics[width=.06\textwidth]{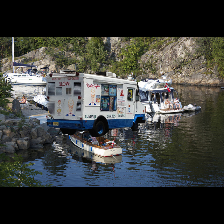} \\
    \includegraphics[width=.06\textwidth]{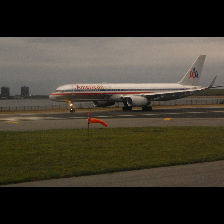} &
    \includegraphics[width=.06\textwidth]{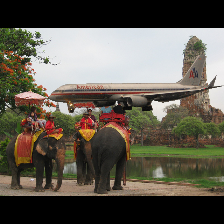} \\
    in-context & out-of-cont. \\
    \multicolumn{2}{c}{(b)} 
  \end{tabular}
  \end{tabular}  
  \caption{{\footnotesize Shifts in conv-$12$ activations on test images (a).
    %
    When the object context contains other objects in
    addition to the isolated object in the first column, we apply the mask
    for these additional objects to the visualizations of the activation
    shifts. Example images in-context and out-of-context (b).}}
    %
%
%
  \label{fig:qualitative}
\end{figure*}

\begin{figure*}[t]
  \centering
  \scriptsize
  \setlength{\tabcolsep}{1pt}
  \renewcommand{\arraystretch}{1}
  \newcolumntype{C}{>{\centering\arraybackslash} m{.09\textwidth}}
  \begin{tabular}{m{.10\textwidth}C|C|C|C|C|C|C||C|C}
    input image &
    \includegraphics[width=.09\textwidth]{images/guided_backprop_new/img208.png} &
    \includegraphics[width=.09\textwidth]{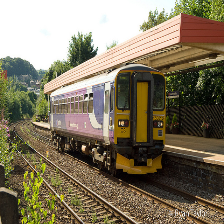} &
    \includegraphics[width=.09\textwidth]{images/guided_backprop_new/img2502.png} &
    \includegraphics[width=.09\textwidth]{images/guided_backprop_new/img10066.png} &
    \includegraphics[width=.09\textwidth]{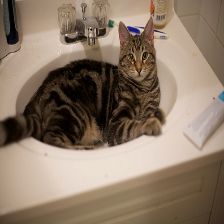} &
    \includegraphics[width=.09\textwidth]{images/guided_backprop_new/img6474.png} &
    \includegraphics[width=.09\textwidth]{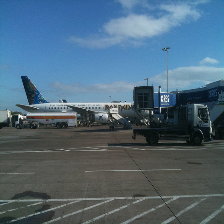} &
    \includegraphics[width=.09\textwidth]{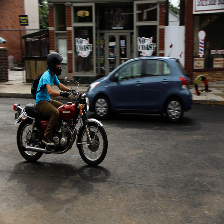} &
    \includegraphics[width=.09\textwidth]{images/guided_backprop_new/img3594.png} \\
    {\bf \compfull (ours)} &
    \includegraphics[width=.09\textwidth]{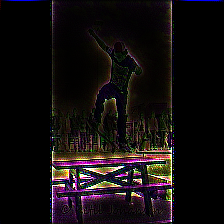} &
    \includegraphics[width=.09\textwidth]{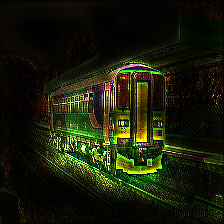} &
    \includegraphics[width=.09\textwidth]{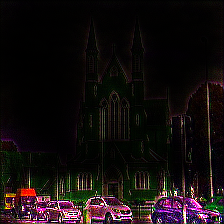} &
    \includegraphics[width=.09\textwidth]{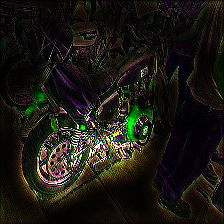} &
    \includegraphics[width=.09\textwidth]{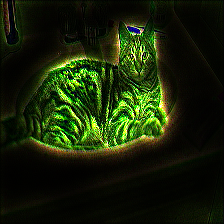} &
    \includegraphics[width=.09\textwidth]{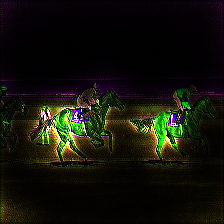} &
    \includegraphics[width=.09\textwidth]{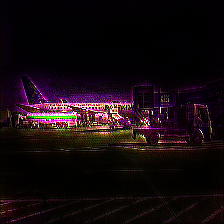} &
    \includegraphics[width=.09\textwidth]{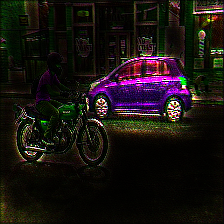} &
    \includegraphics[width=.09\textwidth]{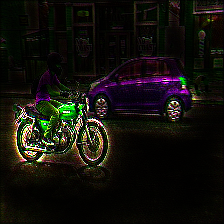} \\                    
    \baselineaug &
    \includegraphics[width=.09\textwidth]{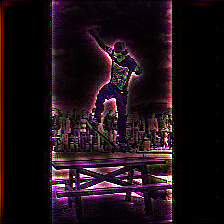} &
    \includegraphics[width=.09\textwidth]{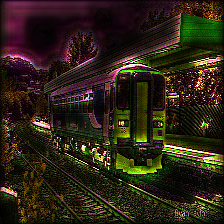} &
    \includegraphics[width=.09\textwidth]{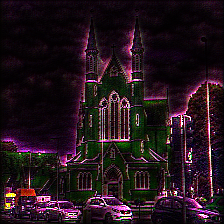} &
    \includegraphics[width=.09\textwidth]{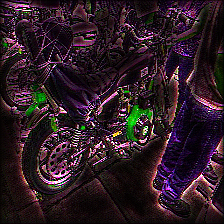} &
    \includegraphics[width=.09\textwidth]{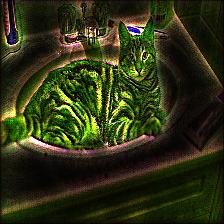} &
    \includegraphics[width=.09\textwidth]{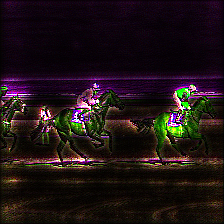} &
    \includegraphics[width=.09\textwidth]{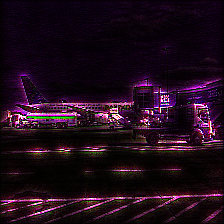} &
    \includegraphics[width=.09\textwidth]{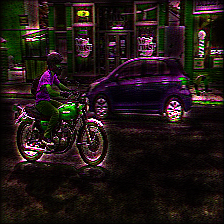} &
    \includegraphics[width=.09\textwidth]{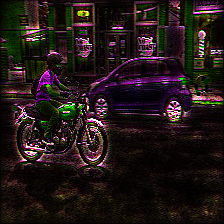} \\
    VGG~\cite{Zisserman15} &
    \includegraphics[width=.09\textwidth]{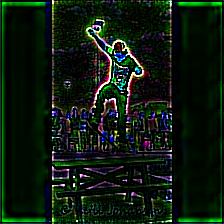} &
    \includegraphics[width=.09\textwidth]{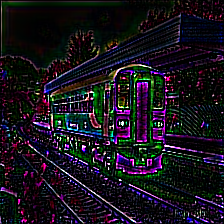} &
    \includegraphics[width=.09\textwidth]{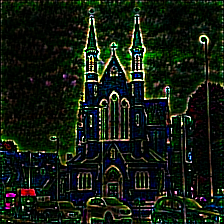} &
    \includegraphics[width=.09\textwidth]{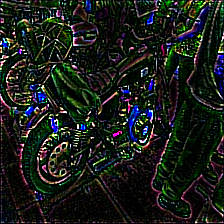} &
    \includegraphics[width=.09\textwidth]{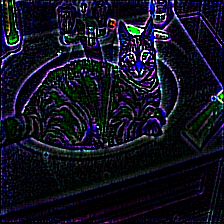} &
    \includegraphics[width=.09\textwidth]{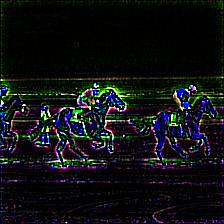} &
    \includegraphics[width=.09\textwidth]{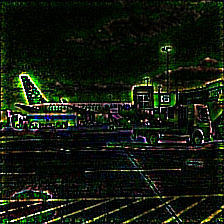} &
    \includegraphics[width=.09\textwidth]{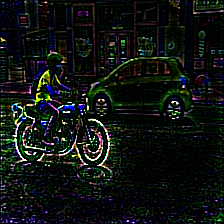} &
    \includegraphics[width=.09\textwidth]{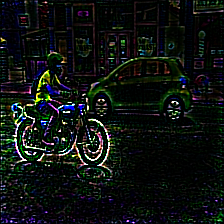} \\
    & bench & train & car & motorcycle & cat & horse & airplane & car & motorcycle \\
    \\
    input image &
    \includegraphics[width=.09\textwidth]{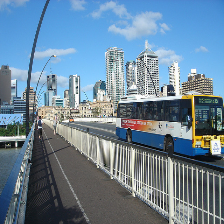} &
    \includegraphics[width=.09\textwidth]{images/guided_backprop_new/img10811.png} &
    \includegraphics[width=.09\textwidth]{images/guided_backprop_new/img4165.png} &
    \includegraphics[width=.09\textwidth]{images/guided_backprop_new/img12096.png} &
    \includegraphics[width=.09\textwidth]{images/guided_backprop_new/img5847.png} &
    \includegraphics[width=.09\textwidth]{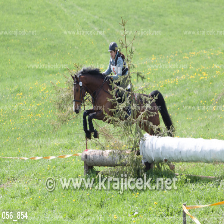} &
    \includegraphics[width=.09\textwidth]{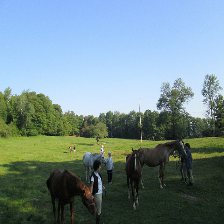} &
    \includegraphics[width=.09\textwidth]{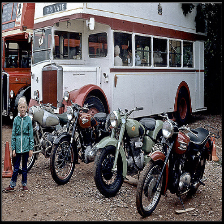} &
    \includegraphics[width=.09\textwidth]{images/guided_backprop_new/img786.png} \\
    {\bf \compfull (ours)} &
    \includegraphics[width=.09\textwidth]{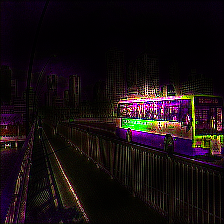} &
    \includegraphics[width=.09\textwidth]{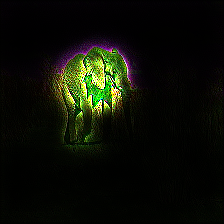} &
    \includegraphics[width=.09\textwidth]{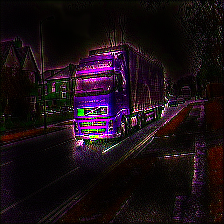} &
    \includegraphics[width=.09\textwidth]{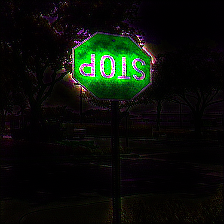} &
    \includegraphics[width=.09\textwidth]{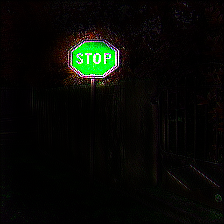} &
    \includegraphics[width=.09\textwidth]{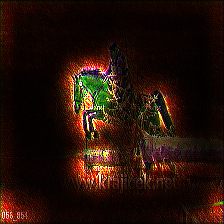} &
    \includegraphics[width=.09\textwidth]{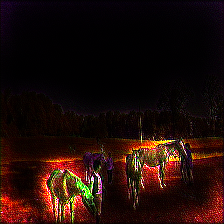} &
    \includegraphics[width=.09\textwidth]{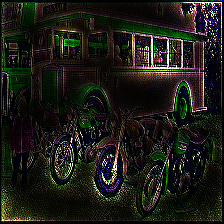} &
    \includegraphics[width=.09\textwidth]{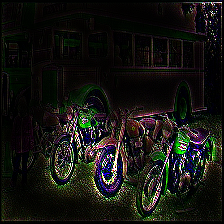} \\
    \baselineaug &
    \includegraphics[width=.09\textwidth]{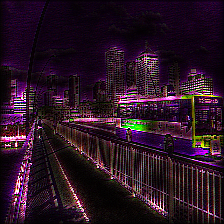} &
    \includegraphics[width=.09\textwidth]{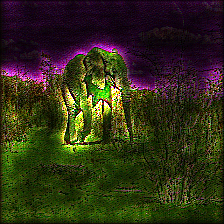} &
    \includegraphics[width=.09\textwidth]{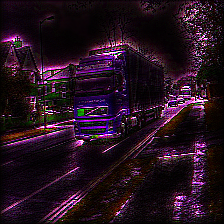} &
    \includegraphics[width=.09\textwidth]{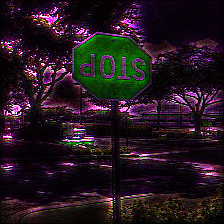} &
    \includegraphics[width=.09\textwidth]{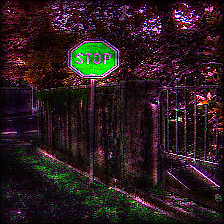} &
    \includegraphics[width=.09\textwidth]{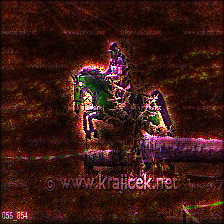} &
    \includegraphics[width=.09\textwidth]{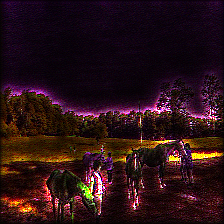} &
    \includegraphics[width=.09\textwidth]{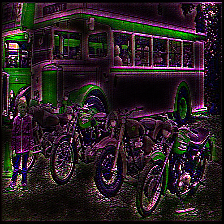} &
    \includegraphics[width=.09\textwidth]{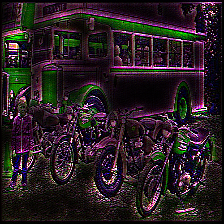} \\
    VGG~\cite{Zisserman15} &
    \includegraphics[width=.09\textwidth]{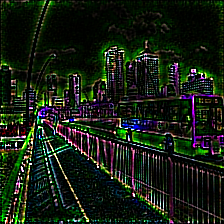} &
    \includegraphics[width=.09\textwidth]{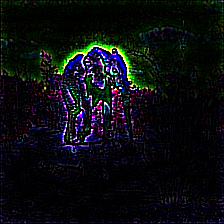} &
    \includegraphics[width=.09\textwidth]{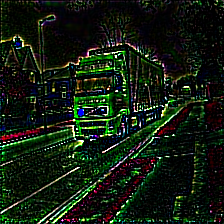} &
    \includegraphics[width=.09\textwidth]{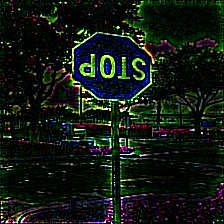} &
    \includegraphics[width=.09\textwidth]{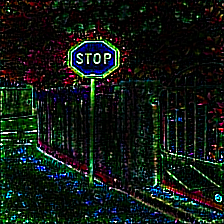} &
    \includegraphics[width=.09\textwidth]{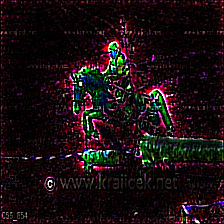} &
    \includegraphics[width=.09\textwidth]{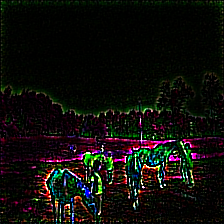} &
    \includegraphics[width=.09\textwidth]{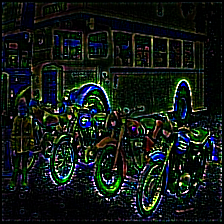} &
    \includegraphics[width=.09\textwidth]{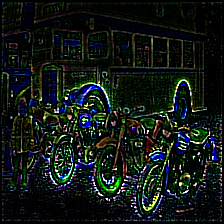} \\    
    & bus & elephant & truck & stop sign & stop sign & horse & horse & bus & motorcycle
  \end{tabular}
  \caption{{\footnotesize Backtracing classification activations (MS-COCO categories, denoted by column
  labels) to test images using
  guided backpropagation~\cite{springenberg15iclrws}.
  Please note the ability of \compfull to backtrace
  to different object categories in one image, whereas \baselineaug
  and VGG produce very similar outputs (rightmost 2 columns). Since VGG was trained on ImageNet categories, which are different from MS-COCO categories, we either backtrace from a semantically close category (identified manually) or VGG's top classification decision when a semantically close category cannot be found.}}
\vspace{-0.4in}

  \label{fig:guided_backprop}
\end{figure*}

Fig.~\ref{fig:mscoco_context} explicitly examines the role of context,
by comparing the performance on the in-
(Fig.~\ref{fig:mscoco_context}~(bottom)) and out-of-context
(Fig.~\ref{fig:mscoco_context}~(top)) test sets defined in
Sect.~\ref{sec:exp_datasets} (Fig.~\ref{fig:qualitative}~(b) gives
examples).
%
Indeed, \compfull improves performance over the
baselines in all cases: 
\compfull outperforms \baselineaugreg by $27.1\%$ (in-context)
and \baselineaugreg by $11.2\%$ (out-of-context).
The relative performance ratio (Tab.~\ref{tab:rel_performance})
between in- and out-of-context objects is slightly more favorable
for \compfull ($0.39$) than for \baselineaugreg ($0.37$).
%
%
%
%
\vspace{-.4cm}
\paragraph{Localization accuracy.}
Fig.~\ref{fig:qualitative} and~\ref{fig:guided_backprop} give
qualitative results that highlight two distinct properties of our
compositional objective \compfull. First, it leads to bottom-up
network activations that are better localized than for \baselineaug,
as indicated visually by the differences in masked and unmasked
activations (Fig.~\ref{fig:qualitative}). 
%
Second, it also leads to better localization when backtracing
classification decisions to the input images, which we implement by
applying guided backpropagation~\cite{springenberg15iclrws}
(Fig.~\ref{fig:guided_backprop}).
Fig.~\ref{tab:compositionality} quantifies this on all test images
of \mscocosub,
by computing the percentage of ``mass'' of the back-trace heat-map
inside the ground-truth mask of the back-traced category, averaged
over categories.
\compfull outperforms both \baselineaug and
VGG~\cite{Zisserman15} by considerable margins.
\vspace{-.4cm}
\paragraph{Discussion.} 
To our knowledge, only~\cite{oquab15cvpr} reports classification (not
detection) performance on \mscoco, achieving $62.8\%$ mAP on the full
set of $80$ classes using fixed lower-layer weights from ImageNet
pre-training~\cite{ImageNetCNN} and an elaborate multi-scale,
sliding-window network architecture.
%
%
In comparison, our \compfull achieves $34\%$ on $20$ classes
(Fig.~\ref{fig:mscoco_sizes}, 'all' column) when trained entirely from
scratch using only a small fraction 
of the full data ($6\%$ with area above $7,000$) 
and a single, fixed scale window (the original image), outperforming
the best baseline \baselinereg by $17\%$.
We believe this to be an encouraging result that is complementary to
the gains reported by~\cite{oquab15cvpr} and leave the combination
of both as a promising avenue for future work.
%
%
%
%
%
%

\vspace{-.2cm}
\section{Conclusion}
\label{sec:conclusion}
%
We have introduced an enhanced CNN architecture and novel loss 
function based on the inductive bias of compositionality. It follows
the intuition that the representation of part of an image should be
similar to the corresponding part of the representation of that image
and is implemented as additional layers and connections of an existing CNN.

Our experiments indicate that the compositionality bias aids
in the learning of representations that generalize better when
training networks from scratch, and improves the performance
in object recognition tasks on both synthetic and real-world data.
Obvious next steps include the application to tasks that explicitly
require spatial localization, such as image
parsing, and combination with pre-trained networks.

\vspace{.1cm}
{\bf Acknowledgments.}
We thank John Bauer and Robert Hafner for support with experiment
infrastructure.
%
%


%
{\small
\bibliographystyle{ieee}
\bibliography{../iclr2015}
}
\end{document}